**REVIEW**

# A Comprehensive Survey on Joint Resource Allocation Strategies in Federated Edge Learning

**Jingbo Zhang[1], Qiong Wu[1,*], Pingyi Fan[2] and Qiang Fan[3]**

[1]School of Internet of Things Engineering, Jiangnan University, Wuxi, 214122, China

[2]Department of Electronic Engineering, Beijing National Research Center for Information Science and Technology, Tsinghua University, Beijing, 100084, China

[3]Qualcomm, San Jose, CA, 95110, USA

*Corresponding Author: Qiong Wu. Email: qiongwu@jiangnan.edu.cn

**ABSTRACT**

Federated Edge Learning (FEL), an emerging distributed Machine Learning (ML) paradigm, enables model training in a distributed environment while ensuring user privacy by using physical separation for each user data. However, with the development of complex application scenarios such as the Internet of Things (IoT) and Smart Earth, the conventional resource allocation schemes can no longer effectively support these growing computational and communication demands. Therefore, joint resource optimization may be the key solution to the scaling problem. This paper simultaneously addresses the multifaceted challenges of computation and communication, with the growing multiple resource demands. We systematically review the joint allocation strategies for different resources (computation, data, communication, and network topology) in FEL, and summarize the advantages in improving system efficiency, reducing latency, enhancing resource utilization and enhancing robustness. In addition, we present the potential ability of joint optimization to enhance privacy preservation by reducing communication requirements, indirectly. This work not only provides theoretical support for resource management in federated learning (FL) systems, but also provides ideas for potential optimal deployment in multiple real-world scenarios. By thoroughly discussing the current challenges and future research directions, it also provides some important insights into multi-resource optimization in complex application environments.

## 1 Introduction

With the proliferation of the Internet of Things (IoT) and mobile devices, the paradigm of data generation and processing is fundamentally changing. In the past decades, data was typically collected and transmitted to remote data centers for processing and analysis. However, with the explosive growth of data volume and the growing concern for low latency, high reliability, and privacy protection, this centralized data processing model is no longer applicable. As a result, edge computing has been proposed as an emerging computing paradigm to address these issues. Edge computing involves performing computing tasks on devices close to the user, thereby reducing the time and network load for data transmission. Currently, Federated learning (FL) is a distributed machine learning (ML) approach that allows different devices to each train models without sharing their data. With the development of technology in recent years, significant progress has been made in theory and practice, and in-depth research has been carried out on model fairness, heterogeneity handling, privacy-preserving security, model training efficiency, robustness, and scalability. The combination of edge computing and the FL will get the benefits from both of them and have more powerful functions in the application. However, FL requires frequent client-to-server communication for updating model parameters, so the communication efficiency of FL is still a key issue. Edge learning (EL) is a method of data processing and model training on edge devices (ED) close to the data source to reduce computing and communication latency [1], improve real-time performance, and reduce the burden on the center server (CS). However, EL usually suffers from resource constraints as well as device heterogeneity. Because of the defective problems of FL and EL, an emerging distributed learning technique called Federated Edge Learning (FEL) has been proposed [2]. In recent years, the development of edge computing technology has provided a solid infrastructure for FEL to better handle the large amount of data generated by IoT devices.

FEL combines the advantages of FL and EL by performing local model computation and training on the ED, which can leverage the computational power of the device for learning, speeding up data processing while preserving data privacy, and then aggregating the updates on the edge server (ES) through the mechanism of FL [3]. The core features of FEL include distributed training, data privacy protection, as well as low latency and high efficiency. FEL achieves distributed training by performing model training on multiple ED, which prevents the transmission of sensitive data to the CS and thus effectively protects user privacy. The approach significantly reduces data transmission latency and improves the overall computational efficiency since data processing and model training occur in the edge network close to the data source. Moreover, FEL is able to utilize distributed computation and storage resources at the edge of the network, thus ensuring effective model training while enhancing data privacy protection [4]. However, the resource allocation problem remains a challenging issue. With the popularity of IoT as well as mobile ED, in the smart planet, a large amount of distributed data is generated, and the combination of FEL and resource allocation is designed to address the resource constraints [5], data privacy, communication efficiency, and energy management in distributed computing environments to achieve efficient model training data processing while protecting data privacy and security. In FEL, resource allocation involves optimizing computational resources, communication resources, network topology, data selection, device scheduling, and energy consumption among participating ED. In fact, optimizing computational resources refers to the process of allocating computational power to the participating ED in FL, and optimizing communication resources refers to the strategy for efficient data transfer and model updating among ED. Optimizing network topology refers to the process of improving system performance and efficiency by adjusting the connections between ED in the network. Optimizing data selection refers to the process of selecting the most valuable data for model training from the data collected from multiple ED. Optimizing device scheduling refers to the process of reasonably allocating and scheduling computing tasks to individual devices in a distributed computing system according to the real-time state and demand of the system, in order to achieve the optimal use of resources and the maximization of system performance. Optimizing energy consumption refers to minimizing the energy consumed by ED during model training and data processing while ensuring system performance, which is crucial for extending the service life of the devices, reducing operating costs and achieving sustainable development. Therefore, an effective resource allocation strategy can not only significantly reduce the communication cost, but also improve the overall system efficiency.

In today's increasingly sophisticated technology, focusing on a single performance metric to the exclusion of other important factors may have limitations to fully deal with complex multi-objective problems, whereas joint optimization can efficiently integrate and weigh different performance metrics in a multi-objective environment by selecting different weight for each metric where multiple available resources can simultaneously get a balance among multiple objectives, satisfying multiple physical constraints. For example, joint optimization of computation and communication resources can balance the weights between computation and communication appropriately, which can significantly improve the overall energy efficiency of the system [6,7]. Joint data selection and communication resources is to select effective data to train the model and address the allocation of communication resources across devices. In such a way, it can effectively protect device privacy, reduce local model training time, and reduce energy consumption [8] to improve learning efficiency [9]. Joint scheduling and communication resource allocation is trying to optimize system performance and efficiency and reduce latency and energy consumption [10,11]. Joint topology and computational resource optimization are usually used in minimizing energy consumption and latency by adjusting the connectivity between ED in the network [12].

In practical applications, joint optimization resources have demonstrated significant benefits. For example, in intelligent transportation systems, joint optimization of computing and communication resources enables real-time adjustment of traffic light control strategies, improving traffic flow and reducing congestion. In smart agriculture, by optimizing data selection and communication resources, edge devices are able to efficiently collect and transmit agricultural data, thereby improving the timeliness of decision-making. In addition, in smart manufacturing, through joint device scheduling and communication resource optimization, the resource utilization and productivity of production equipment is significantly improved. Meanwhile, in financial services, joint optimization of equipment scheduling and communication resources ensures fast processing and security of real-time financial transactions. Similarly, in some smart agriculture, joint optimization of network topology and communication resources improves the efficiency of data collection and transmission in precision agriculture, enabling low-latency agricultural monitoring. The financial domain also improves the collaboration capability of distributed models through joint optimization, which effectively improves the accuracy of fraud detection. Therefore, joint optimization resources can effectively improve system performance in various application scenarios.

This survey bridges the gaps in existing research, mainly from the single resource optimization to the multi-resource co-optimization. This survey will discuss in detail the latest technological developments and cutting-edge advances in FL, with a special focus on joint optimization problems in resource allocation, which is less covered in existing reviews. Through the specific joint optimization case analysis and practical application scenarios, the applications of FEL theory in practice are demonstrated, while some in-depth empirical analysis and future research directions are provided.

This survey will review the research progress in combining FEL with resource allocation, focusing on computational resources, communication resources, data selection, network topology, and device scheduling, and present a deep analysis of the current state-of-the-art methods, as well as point out the existing challenges, and future directions. That is, various resource allocations to enhance the efficiency of optimized FEL and reduce energy consumption will be presented in detail. By synthesizing the existing research results, the advantages as well as limitations will be analyzed, and the existing challenges and future development trends will be discussed.

## 2 FEL Framework

In order to more effectively capture and learn large-scale data from the edge of the network, and provide rapid response and intelligent services for network terminals and other powerful functions, so the edge intelligence technology has been developed. By applying ML algorithms at the edge of the network, it is possible to continuously train and optimize artificial intelligence (AI) models in the edge cloud because ML algorithms can quickly and efficiently utilize distributed mobile data [13]. Among the many branches of edge intelligence technologies, FEL is increasingly becoming one of the most popular EL approaches due to its superior data privacy protection and efficient utilization of endpoint computational resources [4]. FEL is a distributed ML paradigm whose core idea is to perform local data processing and model training on distributed ED [14], while aggregating and protecting user data privacy through physical separation of raw data of users [15]. In FEL, multiple components may be involved, including ED, ES, and potentially cloud servers [16].

### *2.1 FEL Basic Framework and Process*

In FEL, distributed wireless devices collaborate to perform the ML task, and individual ED do not need to upload their complete training data samples to the ES [17]. Instead, each device independently trains the model locally on its local dataset. The model parameters generated as stochastic gradient descent (SGD) are uploaded to the ES. This approach ensures that the original data samples remain on the device locally, thus avoiding the need to share data resources, enhancing data privacy protection. The basic framework of FEL is shown in Fig. 1.

#### *2.1.1 FEL Typical framework*

The typical framework for FEL presented in this paper involves multiple participants, including ED, ES and CS.

1) Edge Devices: refers to devices at the edge of the network, which may include smartphones, sensors, routers, smartwatches, computers, and so on. They have computing and storage capabilities and can perform data processing as well as model training locally.

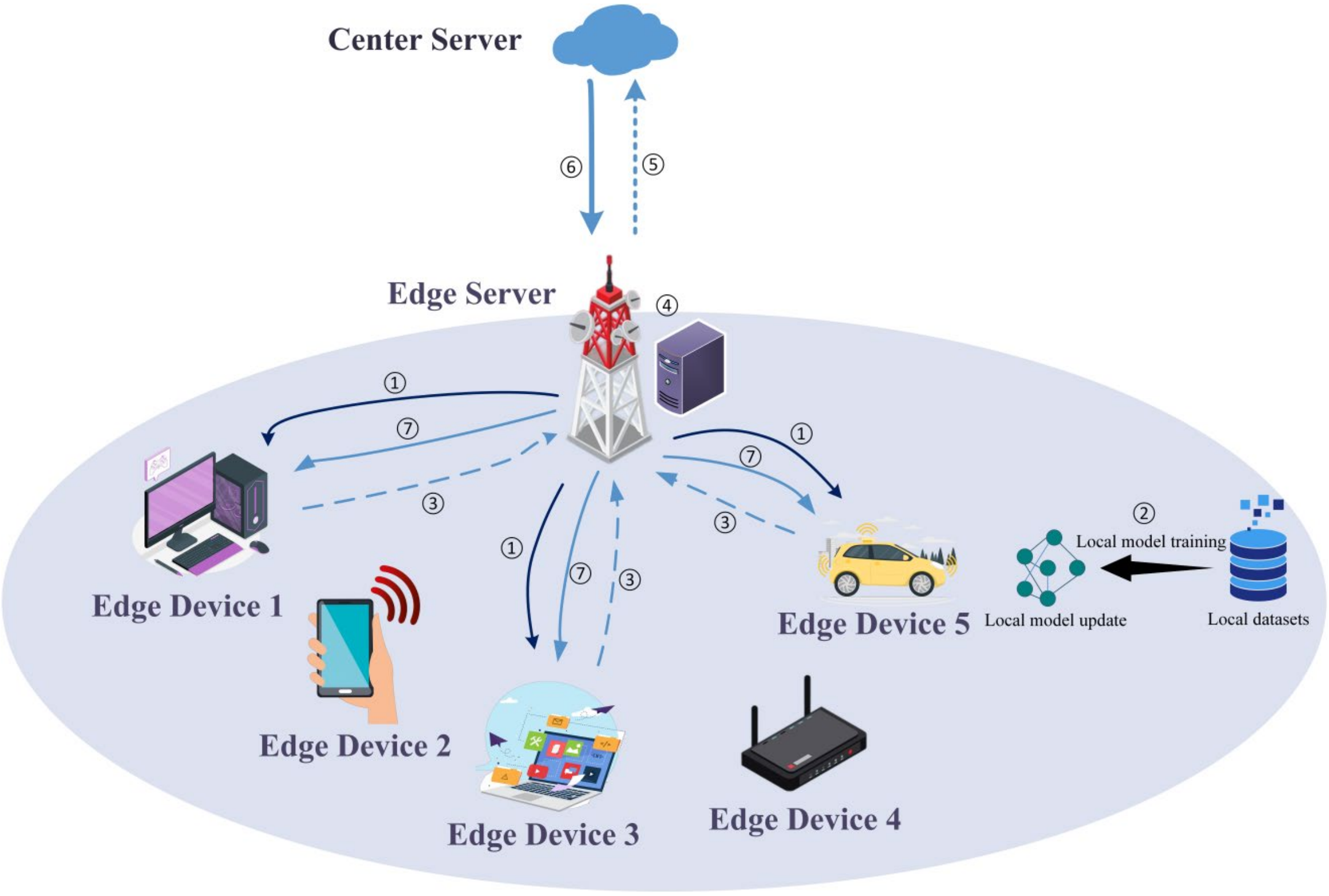

**Figure 1:** FEL basic framework

2) Edge Server: refers to a server that exists at the edge of the network and is responsible for coordinating and managing the computational tasks of the ED. The ES receives model updates from the ED and then aggregates the model updates to finally generate the global model.

3) Cloud server: refers to the CS located in the cloud, which is responsible for coordinating and managing the computational tasks on the ES, and the cloud server receives model updates from the ES, and finally aggregates the model updates to generate the global model.

*2.1.2 FEL Basic Process*

Based on the numbering of the figures in Fig. 1, the basic flow of the FEL can be understood, as shown in Fig. 2.

①Firstly, in the initialization phase, each ED loads its respective initial ML model, and the ES is responsible for coordinating the model training process of the ED and broadcasting the global model.

②In the local training phase, each ED is trained using its own collected data, during which the device updates the local parameters of the model.

③In the model update upload phase, each ED will send the updated model to ES instead of uploading its original data, and ES is responsible for receiving the updated model from each device, which can better protect the user's data privacy.

④In the model aggregation phase, ES aggregates all model updates received to generate a new global model.

⑤In the update phase of the global model, the newly generated global model is sent to the CS, which uses the aggregated updates to compute the global model. To protect data privacy, the CS only receives model updates and does not directly access the raw data.

⑥During the model distribution phase, the updated global model is broadcast to the ES.

⑦The ES is sending the global model back to the relevant ED, and then each ED uses its new received global model and continues its training to update the local model parameters.

The above looping process will continue until the model reaches the expected performance or stops when specific conditions are met. In each iteration, the model is transferred among ED, ES and CS, while the original data is always kept on ED.

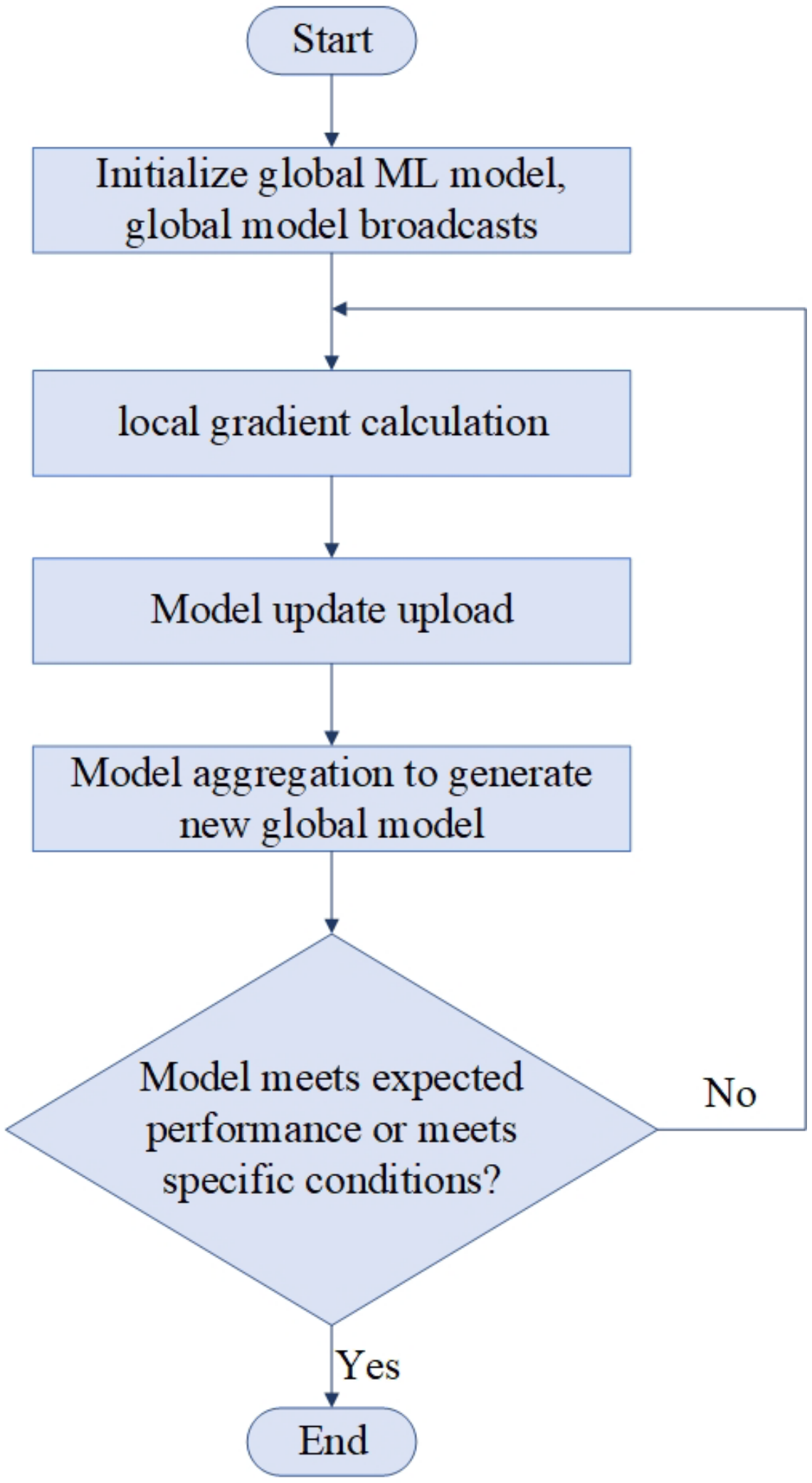

**Figure 2:** FEL basic flowchart local

### *2.2 FEL Key Advantages*

FEL combines the advantages of FL and edge computing to provide an efficient, secure, and expandable approach to data processing and model training. FEL has several advantages over traditional centralized ML and single FL or EL.

#### *2.2.1 Low Latency and Efficient Computing*

In traditional ML, all the data in the device needs to be transmitted to the CS, especially when the data volume is large and widely distributed, which may lead to high latency, however, in FL, the data transmission mainly consists of the model related data transmission, and although some of the data volume is reduced, the model update still needs to be transmitted to the CS for aggregation, which may have latency when the network is unstable or the bandwidth is limited [18]. EL significantly reduces latency by processing data at the ED, which is suitable for applications with high real-time performance [19]. However, the emergence of FEL not only retains the low-latency advantage of EL, which allows real-time processing and response, but also significantly improves the system efficiency through the distributed computation and storage capabilities among the ED. ED processes the data and trains the model locally, which reduces the traffic of model related data transmission to the CS and thus reduces network congestion, can adapt to changes in the local environment in real time, and improves the intelligence of the overall system by sharing knowledge with other devices through the FL mechanism [20].

#### *2.2.2 Data Privacy and Security*

In conventional ML, the data has to be transmitted to the CS for unified processing training, which may have the problem of privacy leakage when some users' private and sensitive data are involved. However, FL performs model training on local devices and shares model updates that are not raw data, which reduces the risk of leaking private raw data [21,22]. However, FL still relies on the CS for aggregation and coordination of model updates [23]. There is still a privacy leakage problem during parameter exchanges in FL [24]. EL in the vicinity of the data source can collaborate on data processing and model training [25], which reduces data transmission and improves data privacy protection, but there are challenges in inter-device collaboration and global model training. Finally, FEL excels in protecting data privacy as it avoids centralized data storage and transmission by training the model locally at the ED and updating it collaboratively with other ED and CS, and the training and updating process does not require the uploading of raw data [26,27]. In addition, ED usually have better security as they are located at the edge of the network and are subject to lower risk of attacks. Performing a large number of computations on ED allows the computational burden of CS and the frequency of data transmission to be further reduced, so that it is evident that FEL is better at protecting privacy compared to others.

#### *2.2.3 Reduce Energy Consumption and Minimize Communication Costs*

In traditional ML, the entire data needs to be uploaded, which results in a large amount of data being transmitted to the CS for processing updates and storage. Therefore, it may cause high communication costs. The CS needs to process a large amount of data from the ED, which requires a large amount of energy consumption. In FL, although there is no more transmission of raw data, frequent model updating and aggregation still consumes a large amount of communication cost [28,29], especially with a large number of participating ED. EL processes the data locally. Requirement for data transmission is reduced, and leads to a reduction in communication costs and energy consumption [30,31]. However, independent ED processing may lead to uneven resource utilization and overall energy consumption is not optimized [32]. However, FEL reduces the amount of raw data transmission to the CS by performing local model training at the ED, so not only the communication cost but also the energy consumption is reduced. The collaborative learning and model update transmission between each ED is relatively small, further minimizing the communication overhead. In addition, the data is distributed and processed locally, which can further optimize the energy consumption of the devices, especially when using low-power devices and optimizing the allocation of computational tasks, making the overall system more efficient and significant in reducing energy consumption [33].

## 3 Related Work to FEL Based Resource Allocation

FEL excels in data privacy protection, real-time processing, and resource optimization. However, due to the heterogeneous nature of ED and limited resources [34], how to efficiently allocate and manage computational and communication resources as well as optimize network topology, data selection, energy consumption, and communication efficiency in the scaling of multiple demands have become important research issues in FEL. In this section, we provide a detailed overview of the current state of research on resource allocation in FEL.

### *3.1 FEL Based Computational Resource Allocation*

Computational resource allocation in FEL involves efficiently distributing computational resources among multiple ED to support the training and updating of distributed machine learning models [35]. These computational resources include central processing units (CPU) , graphics processing units (GPU), memory, storage, and other hardware resources used to perform computational tasks, etc. In FEL, the goal of computational resource allocation is to optimize the efficiency of computational resource utilization and reduce latency while ensuring the quality of model training.

In [36], Zahra et al. utilized the ES available in the environment, which can optimize the total latency and the energy consumption of the IoT devices in the system. Then a Deep Reinforcement Learning (DRL)-based strategy is proposed to decompose the device offloading and resource allocation problem into two independent sub-problems. Specifically, the offloading strategy is adjusted by constantly updating the environment information, and the resource allocation is optimized using the Salp Swarm Algorithm (SSA). Simulation experimental results show that the algorithm achieves significant cost minimization and performs well in terms of latency and power consumption. In [37], in order to minimize the total cost of FL, Liu et al. proposed a DRL based resource allocation algorithm for joint optimization of computational resources and multi-UAV association. With this algorithm, it aims to optimize the allocation of computational resources while reducing the energy consumption of Unmanned Aerial Vehicle (UAV). In order to reduce the computational energy consumption of UAV, a UAV computational resource allocation strategy is proposed. This strategy reduces energy consumption by dynamically adjusting the allocation of computational tasks on the UAV. Ultimately, it is demonstrated that this strategy can effectively reduce the total cost of FL, thus improving the economy and sustainability of the system. In [38], in order to address the computational power limitations encountered by clients when training Deep Neural Networks (DNN) using FL, Wen et al. designed an innovative FL framework by combining edge computing and split learning techniques. This framework allows the model to be split during the training process, which reduces the latency during the training process while ensuring that the test accuracy is maintained. In this way, Wen et al.'s work effectively alleviates the problem of insufficient computational resources on the client side when dealing with complex DNN models, providing a new solution for the application of FL on ED. In [39], Salh et al. develop efficient integration of federated edge smart nodes by investigating the computational resource allocation, optimal transmission power, etc. to reduce the energy consumption of each iteration of the FL time from IoT devices to edge smart nodes and to meet the learning time requirements of all IoT devices. In [40], Zhang et al. designed a FL service-oriented computational allocation policy algorithm to solve the problem of allocation policy computation in time-varying environments. Finally, it is found that the wise allocation of computational resources is crucial for the overall performance of the FL task allocation system. The task allocation is dynamically adjusted according to the real-time computing capability and current load of the device.

In [41], Sardellitti et al. proposed a DRL based task allocation algorithm that is able to adjust the allocation of computational tasks in real-time according to the state of the devices and network conditions. Part of the computational tasks are offloaded to the ES for execution to reduce the computational burden of the ED. In [19], Shi et al. proposed an innovative offloading strategy for edge computing, which achieves optimal allocation of computing resources by subdividing computing tasks into segments and making offloading decisions. Through collaborative computing between devices, this strategy realizes effective offloading and sharing of tasks, and improves the overall computational efficiency and performance of the system. In [42], Liu et al. designed a FL framework that centers on collaborative computing and significantly improves the overall computational efficiency by facilitating the collaboration of tasks among devices. In addition, an energy-aware scheduling method is proposed, which is able to intelligently adjust the allocation of computational tasks by real-time monitoring of the device's battery power and computational load, thus effectively extending the device's endurance. In order to achieve this goal, an energy consumption model is also established, which is able to dynamically adjust the allocation of computing tasks according to the real-time energy consumption of the device, ensuring the effective use of resources and the optimization of energy consumption. In [43], Wang et al. proposed a task scheduling algorithm based on the energy consumption model, which considers a typical EL framework, utilizes limited computational communication resources in EL to the best performance, and optimizes the task scheduling strategy by real-time monitoring of device power consumption. In [44], Zhou et al. used a low-power algorithm to reduce the energy consumption during computation, and designed an energy-efficient FL algorithm through the overall architecture, framework and emerging key technologies for training and reasoning in deep learning models at the edge of the network, which effectively extends the device's endurance by optimizing the energy consumption during the computation process. Through this innovation, the energy efficiency of the FL system is improved, which enables the device to run for a longer period of time while performing learning tasks, thus enhancing the utility and flexibility of the system. For tasks with high real-time requirements, computational resources are prioritized and allocated to improve the real-time response capability of the system. In [45], Li et al. designed a computational scheduling framework centered on real-time prioritization, which significantly improves the real-time response performance of the system by dynamically adjusting the priority of computational tasks. This innovation ensures that the system can respond quickly to real-time tasks and improves the efficiency of the system.

The above studies address a single computational resource allocation, and through various strategies such as task decomposition and scheduling, task offloading, energy-aware scheduling and computational priority scheduling, the utilization of computational resources can be optimized to improve the overall performance and efficiency of the system. However, in [46], computation offloading is a key technique for solving the resource-constrained problem of user devices. Computing on a single device may lead to increased application latency due to limited computation and communication as well as storage resources, computing resources alone may not be sufficient to handle complex and large-scale data streams, leading to processing capacity bottlenecks. On the contrary, for multiple devices, joint scheduling of resource allocation effectively reduces overall latency by optimizing the distribution of tasks across them. In the next chapter, we will explore in-depth in conjunction with other resource allocation algorithms, dynamic scheduling strategies, and co-scheduling mechanisms to jointly enhance the resource management capabilities of FEL.

### *3.2 FEL Based Communication Resource Allocation*

Communication resource allocation for FEL involves efficiently distributing wireless communication resources among multiple ED to support the training and updating of distributed ML model. These communication resources include wireless spectrum, transmission power, bandwidth optimization, and coding resources. In FEL, the goal of communication resource allocation is to optimize the efficiency of communication resource utilization and reduce the latency while ensuring the quality of model training.

FEL faces great challenges in communication resource allocation, especially when the number of ED is huge and the network environment is complex, how to efficiently allocate and manage communication resources becomes a key issue to improve the overall performance of FEL. FEL needs to frequently exchange model parameters between ED and ES, so the optimization of communication bandwidth is especially important. The bandwidth allocation is dynamically adjusted according to the network state and data volume to optimize the data transmission efficiency.

Current research focuses on reducing the amount of transmitted data and improving the transmission efficiency. For example, methods based on model compression and sparsity can significantly reduce the communication overhead [28]. Through data compression and coding techniques, the amount of transmitted data is reduced and the efficiency of communication bandwidth utilization is improved. The size of the model is reduced by removing parameters that contribute less to the model through pruning. In [29], McMahan et al. proposed an SGD algorithm for differential privacy, called Differential Private SGD (DPSGD). This algorithm adds Gaussian noise to the transmitted parameters, thereby reducing the amount of exact data to be transmitted while preserving data privacy. By adding Gaussian noise to the gradient update, the DPSGD algorithm is able to provide effective protection of user privacy without sacrificing model performance. In [47], Lin et al. proposed a gradient quantization technique that effectively reduces the amount of data transmission by reducing the representation accuracy of model parameters. Specifically, the gradient values are converted to a low-precision representation, which significantly reduces the communication cost. With this approach, not only the communication efficiency is improved, but also the network bandwidth requirement is reduced, providing an effective solution for the application of FL systems in resource-constrained environments.

The utilization of communication resources can be effectively improved by dynamically adjusting the communication frequency through real-time monitoring of the network status and device requirements. Based on the current network bandwidth and the computational load of the devices, the communication period of model update can be dynamically adjusted. Based on the real-time network load and data transmission demand, the bandwidth allocation strategy can be dynamically adjusted.

In [43], Wang et al. proposed an adaptive FL mechanism that dynamically adjusts the communication frequency by monitoring the network and device status, thus reducing the communication overhead. This method can flexibly adjust the communication strategy according to the actual conditions, which optimizes the allocation and utilization of communication resources. In [48], Aji et al. significantly reduce the amount of data transmitted by evaluating the importance of gradients, which allows selective transmission of parameters that are critical for model updating. This approach optimizes communication efficiency as it transmits only key information that is critical for model improvement, thus reducing unnecessary data transmission and saving bandwidth and communication resources. In a multi-device environment, a communication prioritization allocation strategy based on device characteristics can be used for more efficient use of communication resources. Each device is assigned a corresponding communication priority based on its computing power, data volume and network condition. This strategy enables the devices to contribute more computational resources and updates during the training process according to their capabilities, which in turn improves the overall model training efficiency. In [49], Nishio et al. proposed that better performing devices can communicate with the server more frequently to provide timely model updates, thus speeding up the training progress. This method of prioritization assignment based on device characteristics helps to achieve collaborative training among multiple devices and improve the overall performance of FL. According to the urgency and importance of tasks, the allocation of communication resources can be flexibly adjusted to prioritize the needs of critical tasks. In [18], Kairouz et al. proposed that more communication bandwidth can be allocated to tasks that are critical for model training or decision making to ensure that they can be completed in a timely and efficient manner. This task-priority based communication resource allocation strategy helps to improve the overall performance of the system, especially when resources are limited, and ensures that critical tasks are prioritized, thus optimizing the response time and resource utilization efficiency of the system.ES plays an important role in FEL by optimizing the resource allocation through auxiliary communication, and ES locally aggregates the model parameters and reduces the communication with the CS. In [19], Shi et al. designed a hierarchical aggregation technique which effectively reduces the overall communication through local aggregation of ES. With this approach, the communication efficiency is significantly improved and the overhead of data transmission is reduced, thus optimizing the system performance. In [30], Mao et al. designed a hierarchical network architecture which optimizes communication resources through multi-layered ES distribution. In this multilayered ES architecture, data can be processed and aggregated at different levels, thus reducing the communication burden on a single node. Communication delay and energy consumption are reduced by optimizing communication paths and selecting the best transmission protocol. Data traffic is decentralized and congestion on a single path is reduced through multi-path transmission techniques. It also improves communication efficiency and reliability by selecting the optimal transmission protocol according to the network conditions.

The communication resource allocation based on FEL effectively optimizes the communication bandwidth utilization through the strategies of model compression, adaptive communication, inter-device collaboration and ES assistance. Separate communication resource allocation had been extensively studied in the aforementioned works, aiming to optimize the efficiency of federated learning through effective communication resource management. In contrast, federated resource allocation approaches can bring more performance improvements. In [50], Ni et al. suggested that in complex tasks with multiple rounds of updates, leading to high communication costs, increasing network bottlenecks and affecting system performance. In addition, load imbalance is a serious problem and may lead to degradation of quality of service. Joint resource allocation addresses these challenges more effectively by sharing the load and improving system reliability and flexibility. Joint resource allocation not only enhances the flexibility and robustness of the FEL system compared to single communication resource optimization, but also substantially improves the model performance in multi-device and multi-task environments, making it more efficient and practical in real-world applications. Future research can be combined with other systematic processing methods to jointly enhance the resource management capability of FEL.

### *3.3 FEL Based Data Selection*

Due to the uneven and diverse data distribution on each device, the role of data selection strategies in FEL is crucial. This subsection explores effective data selection strategies in the FEL framework to enhance model performance and training efficiency.

In FEL, data selection not only affects the performance of the model, but also directly relates to the efficiency of training and privacy protection. Data on different ED may have significant differences, i.e., data heterogeneity, and such differences can lead to inconsistency in model performance. If an effective data selection strategy is not chosen during model training, it may result in data on certain devices contributing less to the model, thus affecting the overall model accuracy and generalization ability. The privacy of the devices should also be protected during the data selection process while maintaining the validity and availability of the data.

In [51], Serhani et al. designed a dynamic sample selection model designed to optimize resource utilization and address the problem of big data heterogeneity and data imbalance. On ED, computational and storage resources are usually limited. Therefore, by selecting a subset that best represents the overall data distribution for training, resource utilization efficiency can be significantly improved and unnecessary computational overhead can be reduced. In [52], Hu et al. developed a framework called Auto FL, which enhances device involvement in model training by empowering clients to make autonomous decisions. This autonomous decision-making process allows the client to flexibly decide whether or not to participate in model training based on its own resource status and network conditions, thus effectively increasing the device's participation. This increased participation not only expands the range of devices participating in training, but also significantly improves the performance and training efficiency of FEL. In FEL, data selection also requires consideration of privacy protection. By selecting the non-sensitive data that contributes the most to the model training, the model performance can be improved while ensuring privacy. In [53], Wei et al. proposed a participant selection problem aiming to minimize the total cost of hierarchical FL with multiple models, and based on this, to enhance privacy protection and reduce the total learning cost. In [54], a wireless gradient aggregation technique is employed in order to achieve efficient resource management in FEL. In order to solve the channel and data distortion problem caused by channel fading and data-aware scheduling, Su et al. proposed a sensor-side residual feedback mechanism. This mechanism is able to offset the bias caused by channel and data distortion, which improves the convergence speed of training, reduces the training loss, and avoids the distortion problem. In this way, not only faster convergence speed and lower training loss are realized, but also the accuracy and reliability of the training process are ensured. In [55], Kim et al. developed an innovative mobile edge computing (MEC) server selection and datasets management mechanism for FL-based mobile network traffic prediction. The impact of MEC server participation and datasets utilization on global model accuracy and training cost is deeply analyzed to construct an accurate mixed-integer nonlinear programming problem. Validated by simulation experiments and real datasets, the proposed framework achieves a 40% reduction in energy consumption while reducing the number of MEC servers involved in the FL process and maintains a high prediction accuracy.

Data selection plays a crucial role in FEL. An effective data selection strategy not only improves the performance of the model, but also optimizes resource utilization, improves training efficiency and protects data privacy. However, separate data selection strategies, while optimizing the quality of training data to some extent, often fail to take full advantage of their inherent values when communication bandwidth is limited or computational power is restricted. In [56], Xin et al.

suggested that a single data selection strategy has several drawbacks. It can lead to poor data representation, significant model bias, high risk of over-fitting, and poor performance in dealing with data imbalance and insufficient update frequency. Comparatively, joint resource allocation, by combining data sources from multiple nodes, can provide more comprehensive data diversity, reduce bias, balance data categories, and improve the generalization ability and updating effectiveness of the model. Future research can further explore more data selection methods and combine them with other resources (computational resources, communication resources). Compared to individual data selection, the joint resource allocation approach not only considers the importance of data selection, but also simultaneously optimizes multiple dimensions such as communication resources, computational resources, and device scheduling. This integrated optimization approach can better coordinate the use of data selection with other system resources, enabling the system to achieve better performance under different network conditions and computing environments when facing the growing multiple demands of users.

### *3.4 FEL Based Device Scheduling*

In FEL, the heterogeneity of ED makes device scheduling crucial. A reasonable device scheduling strategy can fully utilize the computational resources of ED, balance the load, reduce the training time, and improve the overall performance of the system. The wide variation in ED in terms of computational power, storage capacity, network bandwidth, and battery life pose challenges to device scheduling. How to maximize the utilization of these heterogeneous resources while ensuring the model training effect is the core problem of device scheduling. The main goals of scheduling strategies include maximizing resource utilization, balancing device load, minimizing model training time, optimizing communication overhead, and extending device battery life, etc.

Resources such as computing power, storage space, and energy supply of the ED are evaluated to determine the performance and capacity of the equipment. Since the ED has limited resources, the device capacity assessment can optimize the resource utilization and improve the scalability and utility of the system.

Through the application of DRL algorithm, it can dynamically learn and adjust the task allocation strategy according to the real-time state of the equipment and network conditions, so as to improve the efficiency of resource utilization. In [57], Chen et al. constructed an adaptive scheduling model for energy hubs based on the DRL algorithm. The model adopts the federated DRL method to meet the requirements of data privacy protection. The matching learning technique accelerates the convergence of DRL intelligence by solving the training instability problem in large-scale shared learning. The algorithm significantly improves the training efficiency and economic benefits in the application of multi-energy hubs. In [58], Yan et al. pointed out that due to the rapid development of IoT and edge computing technologies in recent years, personal privacy and data leakage have become major issues in IoT edge computing environments. FL has been proposed as a solution to address these privacy issues. However, the heterogeneity of the devices in the IoT edge computing environments poses a significant challenge to the implementation of FL. To overcome this challenge, a novel DRL based node selection strategy is proposed to optimize FL in heterogeneous device IoT environments. Furthermore, the proposed strategy can ensure the efficiency of heterogeneous devices participating in the training and improve the accuracy of the model while guaranteeing privacy protection.

Dynamic task allocation strategies achieve efficient resource utilization by dynamically adjusting the allocation of tasks based on the real-time state of the device. In [59], Ren et al. introduced an innovative scheduling strategy aimed at improving the performance of FEL by developing a probabilistic-based scheduling framework that ensures unbiased aggregation of global gradients and accelerates the convergence speed of the model. The strategy designs a relatively optimal scheduling scheme by taking into account the importance of channel states and update data. In addition, a detailed convergence analysis is performed to demonstrate the effectiveness of the strategy. The final experimental results show that the strategy can significantly improve the learning performance of FEL. In [60], Chu et al. explored the Constrained Markov Decision Process (CMDP) problem of combining FL with MEC server. In this model, the mobile device periodically transmits updates of the local model to the ES, which contains training on locally sensitive data. ES is responsible for aggregating the parameters from the mobile device and broadcasting the aggregated parameters back to the mobile device. The ultimate goal is to dynamically optimize the mobile device's transmit power and the scheduling of training tasks. In the first step, the resource scheduling problem during synchronized FL is modeled as a CMDP problem and the size of the training samples is used as a measure of FL performance. Due to the coupling between iterations and the complexity of the state-action space, the authors employ a Lagrange multiplier approach to solve this problem. The final simulation results show that the proposed stochastic learning algorithm outperforms other benchmark algorithms in terms of performance. This suggests that the strategy can effectively balance the performance and resource constraints of FL, thus realizing efficient FL in a MEC environment.

Energy-aware scheduling strategies dynamically adjust task assignments to reduce energy consumption through the real-time energy consumption of devices. In [61], Hu et al. conducted an in-depth study on the problem of device scheduling in FEL systems, which face stochastic data generation constraints of energy and delay at the ED. To cope with the system dynamics of data arrivals and energy consumption, a dynamic scheduling algorithm was designed using Liapunov optimization, which aims to maximize the importance of long-term data while taking into account the energy consumption and per-round latency under the constraints of the set of scheduling devices. The final results show that the proposed method has significant advantages in reducing energy consumption and achieving better learning performance. This suggests that the strategy can effectively balance the performance and resource constraints of the FEL system to achieve efficient FEL in practical applications. Both energy consumption and model performance are important metrics of FEL. In [62], Hu et al. model the two metrics, energy consumption and model performance, to reveal the relationship between them, especially the correlation with the size of training data. In order to further optimize the FEL system, a workload constraint is added to the model, resulting in a common factor constraint problem. A strategy for resource optimization and device scheduling is proposed to address this problem, aiming to achieve a balance between energy consumption and model performance. The strategy minimizes the energy consumption of the training device by dynamically adjusting the workload of the device and the size of the training data. Experimental results demonstrate the effectiveness of the strategy, which can significantly reduce the energy consumption of the training device while maintaining good model performance.

In FEL, communication time minimization is necessary for device scheduling. In FEL, due to bandwidth constraints, only some devices can be selected to upload their model updates at each training iteration. This challenge has led to the study of optimal device scheduling strategies in FEL aiming at minimizing the communication time. In [11], Zhang et al. proposed a new probabilistic scheduling scheme aimed at minimizing the communication time. The final results of this scheme show that the method is robust to changes in different network conditions and device performances, and it is effective in minimizing the communication time. In [63], Zhang et al. focused on exploring the optimal device scheduling strategy in FEL to significantly reduce the communication time. However, due to the difficulty in quantifying the communication time, the current study can only partially address this issue by considering the number of communication rounds or the delay per round to indirectly determine the total communication time. In order to address this challenge more precisely, a first attempt is made to formulate and solve the communication time minimization problem. Based on the analytical results, a closed form of an optimized probabilistic scheduling policy is obtained by solving an approximate communication time minimization problem. As the training process progresses, this optimized strategy gradually shifts the priority from reducing the number of remaining communication rounds to reducing the delay per round. Ultimately, the proposed scheme is demonstrated to be effective in minimizing the communication time and reducing the delay per round in a case study of cooperative 3D target detection.

FEL based device scheduling effectively improves the overall system performance and resource utilization efficiency through the application of DRL algorithms, dynamic task allocation, energy-aware scheduling, and minimization of communication time. Unlike standalone device scheduling, joint device scheduling and communication resource allocation not only focuses on selecting which devices participate in training, but also simultaneously optimizes each device's communication resources, bandwidth allocation, and computational capabilities. In [64], Taïk et al. suggested that single-device scheduling strategies face problems such as poor data representation, uneven resource utilization, large fluctuations in model performance, data imbalance, and insufficient update frequency. In contrast, joint resource allocation can improve data diversity, optimize resource allocation, enhance model stability, balance data categories, and increase update frequency and efficiency by integrating data from multiple devices. Future research can delve into more intelligent scheduling algorithms, real-time dynamic adjustment strategies, and multi-level collaborative mechanisms. By jointly optimizing additional resources, FEL systems can better coordinate the relationship between device selection and other resources allocation.

### *3.5 FEL Based Network Topology*

Network topology is the structure of devices and their connections in a network. In FEL, network topology directly affects data transmission, model synchronization and resource allocation. FEL based network topology is a distributed computing architecture that combines edge computing and FL to enable collaborative learning among devices while protecting data privacy and reducing communication overhead. In this topology, ED such as smartphones, sensors, and IoT devices collect data locally and train on local models. These devices then share updates to the model, typically usually gradient or parameter updates, rather than raw data, in a privacy-preserving manner, either through ES or by directly collaborating with other devices.

Network topology in FEL can be categorized as star topology, tree topology, mesh topology, and hybrid topology, etc.

The origin of FL is the co-training of ML algorithms on devices in multiple locations, and the core idea is to implement ML without centralizing or directly exchanging private user data. Nevertheless, most FL implementations still rely on the presence of CS. The most common network topology in FL, including the initial work, uses a centralized aggregation and distribution architecture, which is also known as a star topology. Thus, the graphical representation of this server-client architecture resembles a star, as shown in Fig. 3. Many FL studies and algorithms have been designed based on the assumption of this star topology. In a star topology, all ED are directly connected to a CS, which is responsible for coordinating the aggregation and distribution of models. The advantage of this topology is that the CS can easily manage all the ED, but the disadvantage is that the CS then becomes a single point of failure and can lead to bottlenecks in large-scale networks. In [65], Wu et al. consider that traditional FedAvg based FL approaches tend to use a simple star network structure, which does not adequately consider the changes in edge computing environments in real-world scenarios and the diverse heterogeneous and hierarchical characteristics of the network topology. Therefore, other topology needs to be considered to address the limitations and bottlenecks of the star topology.

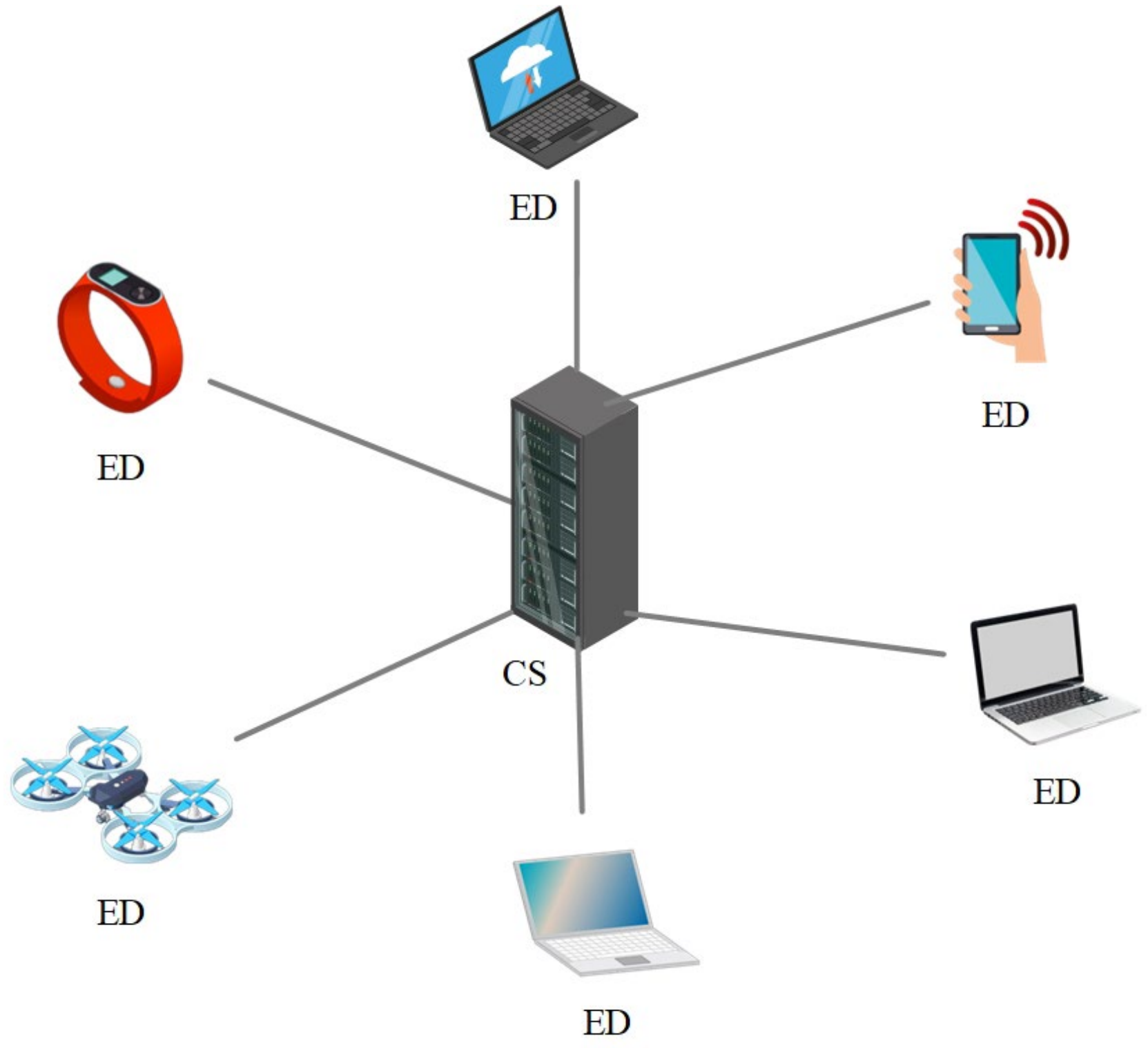

**Figure 3:** star topology

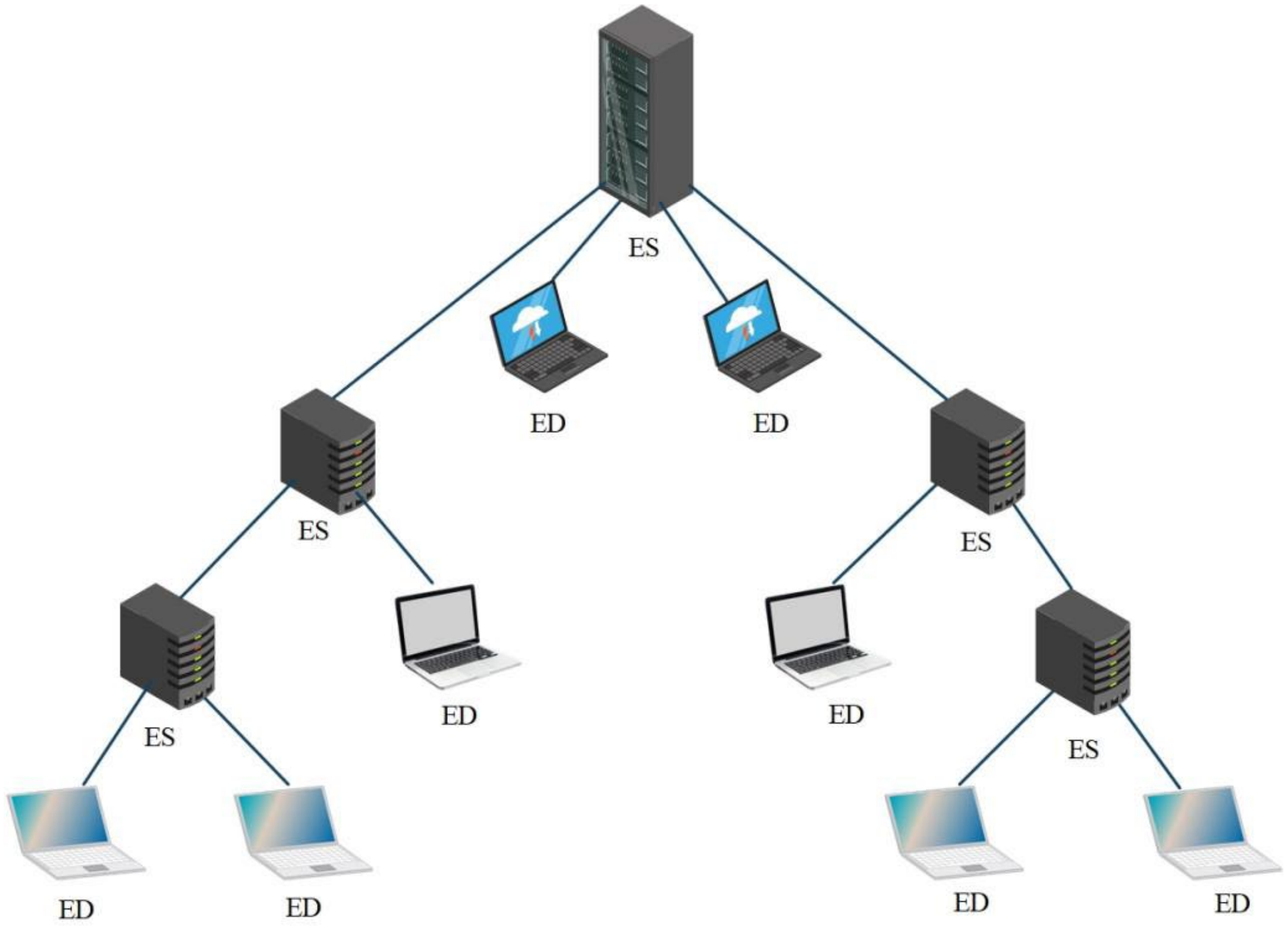

**Figure 4:** tree topology

Between the CS and the ED, one or more additional levels may exist. For example, the ES connecting the ED and the CS may form one or more layers to form a tree topology, as shown in Fig. 4. In this structure, the CS is located at the top of the tree and the ED is located at the bottom of the tree. A tree topology contains at least three levels, otherwise it would be considered as a star topology. Tree topology helps to overcome the bottleneck in performance and single point of failure in star topology as compared to conventional FL. Thus, FL with tree topology can achieve different communication costs in different clusters based on their energy profiles. In the FL system with tree topology, any level of clients can exist. Fuzzy logic research in tree topology is divided into two main categories: hierarchical and dynamic. Hierarchical research is concerned with how to rationally divide and design the hierarchical structure of the tree topology in order to improve the overall performance and stability of the system. Dynamic research, on the other hand, focuses on how to dynamically adjust the hierarchical structure of the tree topology according to the changes in the system state and requirements in order to adapt to different application scenarios and environmental conditions. The tree topology is similar to a hierarchical network, where ED are connected according to a hierarchical structure and data and models can be transmitted along tree paths. This structure reduces the risk of single point of failure and can be scaled efficiently, but bottlenecks may still exist at some levels. In [66], Landola et al. In using a tree structure for data or gradient information transfer, information is aggregated at the higher nodes of each level. This approach not only scales down the amount of data transmitted, but also effectively alleviates network congestion. Compared with the traditional parameter server architecture, the synchronization model requires less bandwidth during data transmission.

A mesh topology is a network structure in which all end devices are directly connected in a local network. In recent research, this topology is often used in FL systems. For example, Peer-to-Peer (P2P) or Device-to-Device (D2D) FL models are mesh topology. Nevertheless, many existing FL systems still rely on centralized or cloud servers for model aggregation. When centralized servers are not feasible, decentralized approaches are sometimes considered as a sub-optimal alternative to centralized approaches. In a mesh topology, each ED can be directly connected to multiple other devices to form a mesh-like network, as shown in Fig. 5. The mesh topology provides high redundancy and fault tolerance, but the network complexity is higher and more difficult to manage and maintain. In [67], Xu et al. proposed a decentralized learning architecture using a mesh topology with multiple autonomous computing nodes. In this setup, all nodes are equal and communicate directly with each other through P2P, promoting efficient and flexible data exchange.

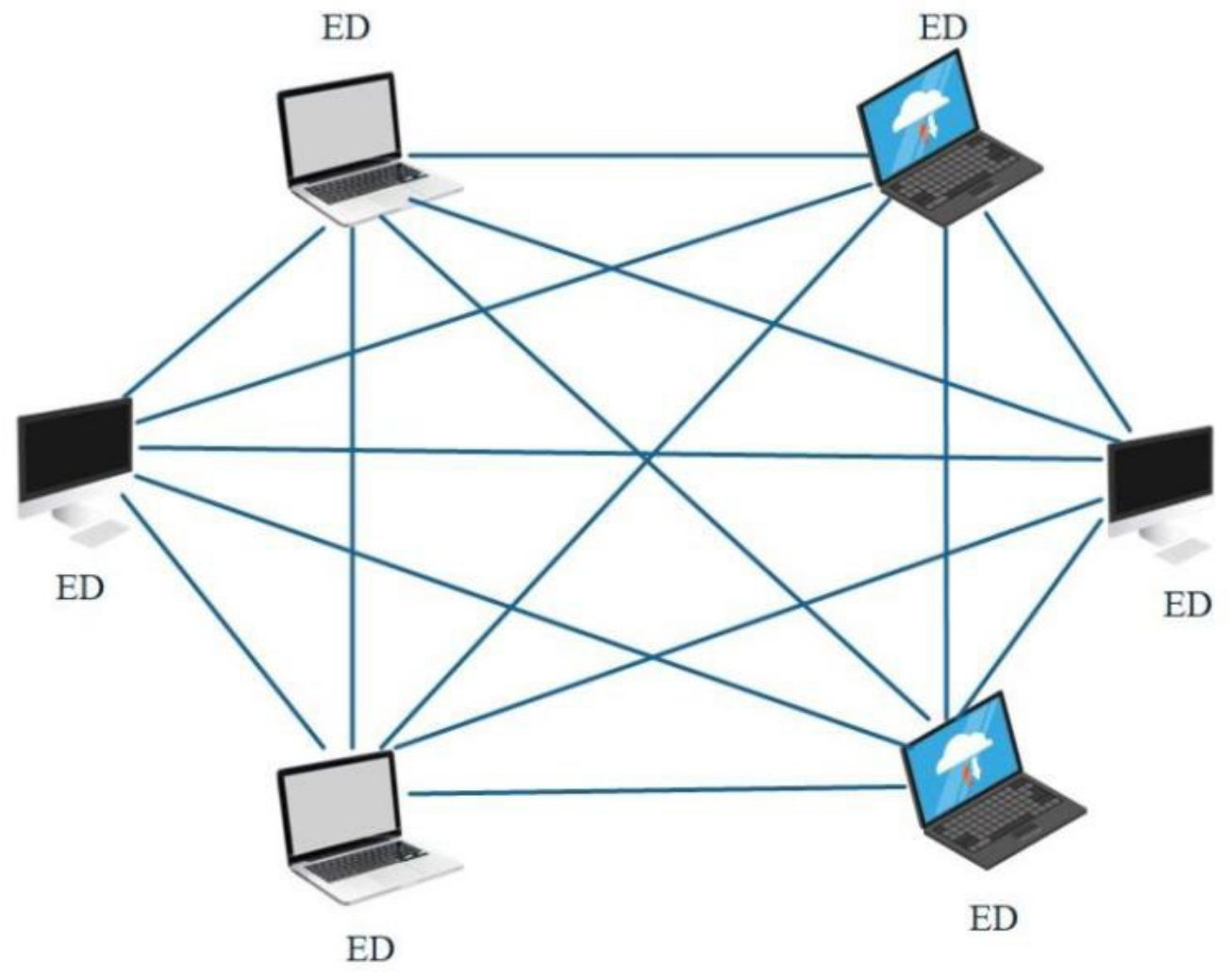

**Figure 5:** mesh topology

In the previous study, we explored a series of topology that are prevalent in the FL domain. Although these topologies are able to cope with numerous scenarios, each of them has certain advantages and limitations. In order to fully utilize the advantages of different topology, attempts have been made to fuse various topology to form a hybrid network topology. The design concept of this hybrid topology is to combine the advantages of at least two traditional architectures in order to seek a balance between performance and complexity, and thus create a more flexible and efficient solution, as shown in Fig. 6, where the addition of mesh connections can be added on top of a star, and the redundancy and fault tolerance can be improved. In [68], Kaur et al. modeled and simulated a wireless sensor network using Zigbee technology according to the characteristics of the Zigbee protocol, designed a hybrid topology, considered three unused combinations of possibilities of routing schemes for Zigbee in different scenarios, and finally verified the communication by testing key metrics such as latency, throughput, network loading, and packet delivery rate, and finally verified the communication reliability of the network.

Network topology plays a crucial role in FEL. Properly designing the network topology and adopting some dynamic and collaborative resource allocation strategies can significantly improve the performance and efficiency of FEL as well as the stability of the system. However, under complex network conditions, it is difficult to achieve globally optimal performance with separate topology optimization strategies. In contrast, the joint resource allocation strategy is able to optimize the network topology and communication resources together as a whole. In [69], Wei et al. noted the limitations of single network topology resource allocation, including constraints and high communication costs, and stressed that federated resource allocation improves flexibility and scalability. Under the framework of joint network topology and communication resource allocation, not only can the network topology be dynamically adjusted according to the geographic location of devices, network structure, and communication status between devices, but also the allocation of communication resources can be optimized simultaneously. This strategy can effectively reduce the communication delay under different topologies and maximize the participation of devices and the overall efficiency of the system. The next chapter will further explore more ways to combine network topology with other resource allocations.

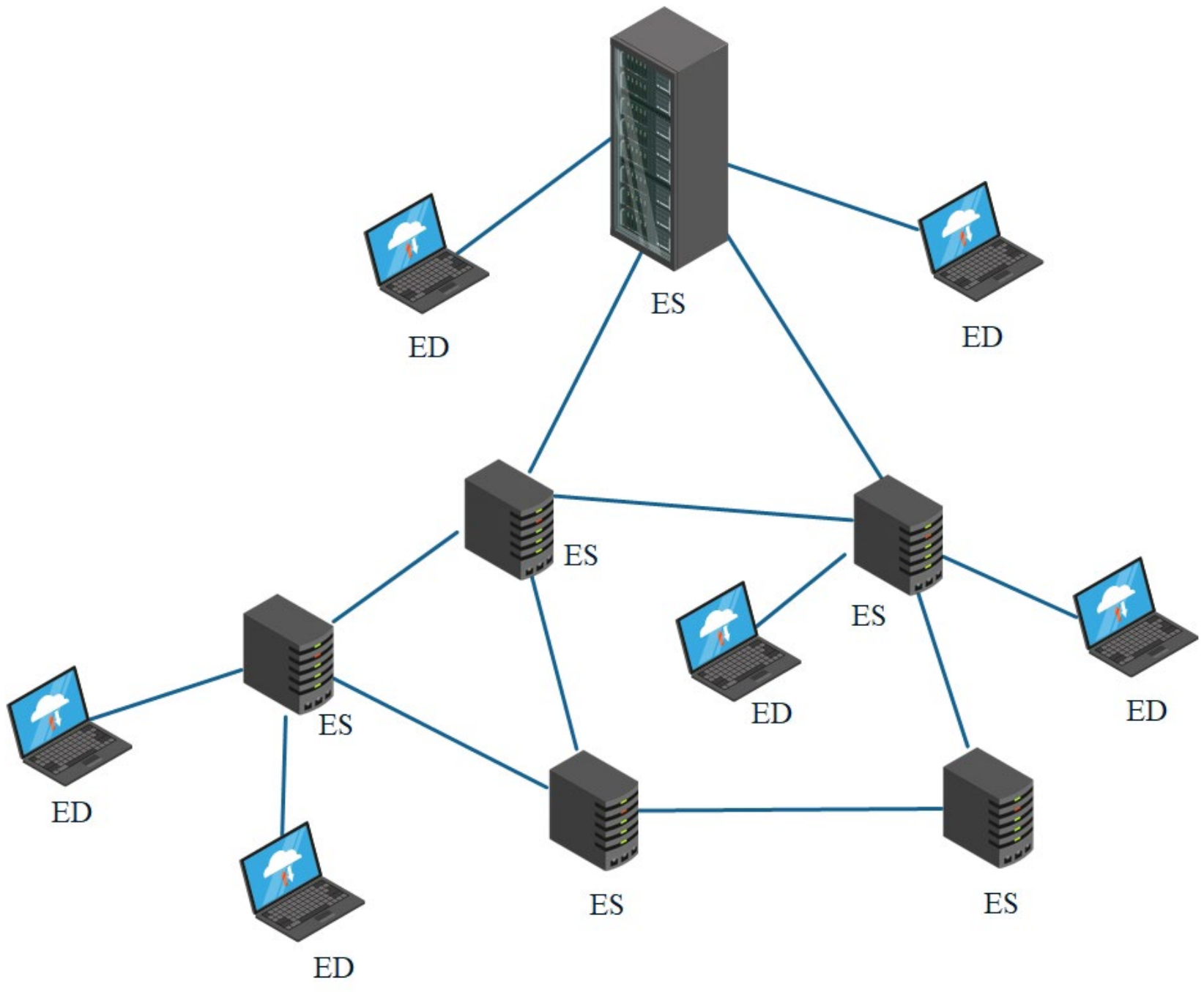

**Figure 6:** hybrid topology

## 4 Optimized Multi-Resource Allocation for FEL

With the popularity of IoT devices and the explosive growth of data volume, the traditional centralized data processing paradigm can no longer meet the demands of real-time and privacy protection. Therefore, FEL, an emerging distributed ML paradigm, has been proposed to address these issues. FEL allows model training and updating on distributed devices at the edge of the network, thereby reducing data transmission latency, improving computational efficiency, and enhancing privacy protection. However, FEL faces many challenges in its implementation, one of which is how to efficiently allocate and manage multiple resources to ensure model training efficiency and performance. In FEL, multi-resource allocation involves how to rationally allocate computational resources, communication resources, data selection, device scheduling, and network topology to support model training and updating. In this section, we explore the issue of multi-resource allocation in FEL.

### *4.1 Joint Computing and Communication Resource Optimization for FEL*

In FEL, computational and communication resources are closely related. the computational capacity of ED determines the speed and energy consumption of local model training, while the limitation of communication resources affects the transmission efficiency of model parameters. Jointly optimizing the computational and communication resources can reduce the training time and energy consumption and improve the overall efficiency of the system under the premise of ensuring the model accuracy. Existing research focuses on the following aspects, dynamic task allocation and scheduling, model compression and communication optimization, and energy consumption awareness and energy saving optimization.

#### *4.1.1 Dynamic Tasking and Scheduling*

Dynamic task allocation and scheduling is to dynamically adjust the allocation and communication frequency of computing tasks according to the real-time status of ED. Dynamic task allocation and scheduling contains not only task decomposition and dynamic scheduling, but also load balancing and computational prioritization. Among them, the task decomposition and dynamic scheduling technique improves the efficiency of resource utilization, improves the training efficiency, and reduces the overall training time while fully utilizing the computational resources of the ED by decomposing the computational task into a number of sub-tasks, decomposing the model into a number of sub-models, and dynamically adjusting the order of the task execution allocation and scheduling according to the real-time computational capacity of the device system and the network condition [70]. Task decomposition is the process of decomposing a large-scale model training task is decomposed into multiple small tasks and assigned to different ED.

In [45], Li et al. proposed a task decomposition method based on ED collaboration, which optimizes resource utilization by decomposing and assigning tasks. Dynamic scheduling refers to the dynamic adjustment of task allocation and scheduling based on the real-time computing capability, load and network conditions of the device, then in [71], Chen et al. proposed a DRL based dynamic scheduling algorithm to achieve efficient resource allocation by monitoring and predicting the device status in real-time. Load Balancing and Computation Prioritization in Dynamic Task Allocation and Scheduling Through load balancing and computation priority setting, it ensures that the computational load of each ED is balanced, avoids overloading of a single device, and effectively improves the computational efficiency of the overall system. Load balancing refers to real-time monitoring of the computing load of each device according to the computing capacity, storage space and energy supply and other resources of the ED, and reasonably allocating tasks, and realizing load balancing by adjusting the task allocation, and ultimately achieving the optimal use of resources. Among them, load balancing can improve the overall performance of the system and avoid resource waste and bottleneck problems. In [72], Zhao et al. proposed a load balancing framework to optimize resource utilization by dynamically adjusting task allocation through distributed algorithms. Computational prioritization refers to setting the computational priority according to the importance and urgency of tasks and prioritizing the allocation of resources to high-priority tasks. Computational prioritization ensures that critical tasks are prioritized and improves the response speed and performance of the system. In [49], Nishio et al. proposed a priority scheduling algorithm that improves system response speed by prioritizing high priority tasks. In [73], Gu et al. proposed an improved federated self-supervised learning algorithm. It optimizes computational and communication resource allocation by combining integrated sensing and communication technologies in an intelligent transport system. The algorithm does this by offloading some tasks to the roadside unit (RSU). It also adjusts the transmission power, CPU frequency and task allocation ratio to balance the energy efficiency of local computing with RSU computing while optimizing resource allocation. The study shows that the improved algorithm reduces energy consumption and improves offloading efficiency, demonstrating its effectiveness in dynamic task offloading and resource allocation.

In FEL, dynamic task allocation and scheduling is a complex process that requires comprehensive consideration and optimization of factors such as task decomposition, dynamic scheduling, load balancing, and computational priority. By reasonably allocating tasks and scheduling, the overall performance and efficiency of the system can be improved and the optimal utilization of resources can be achieved.

#### *4.1.2 Model Compression and Communication Optimization*

Model compression and communication optimization are used to reduce the amount of transmitted data and optimize communication resources through model compression and parameter quantization [28]. Model compression techniques significantly reduce the amount of transmitted data, reduce storage and computation requirements, and improve the efficiency of communication bandwidth utilization by reducing the number of model parameters, using simpler model architectures, and so on [47]. In FEL, since ED are usually limited in terms of resources, such as computational power, storage space, and energy supply, model compression is crucial to improve the scalability and utility of the system. Parameter pruning in model compression techniques is used to reduce the size of the model by removing parameters that contribute less to the model through pruning.

A weight pruning method significantly reduces the number of parameters of the DNN by pruning and fine-tuning layer by layer to achieve model compression. The low-rank decomposition in the model compression technique is to reduce the number of parameters by decomposing the weight matrix into two smaller matrices. This approach reduces the model storage space and improves the computational efficiency. Communication Optimization Strategies In FEL, the performance of the system is optimized by reducing the unnecessary communication overhead, optimizing the communication resources, improving the communication efficiency and reducing the communication delay through strategies such as adaptive communication frequency and differential privacy. In FEL, communication overhead is an important performance bottleneck. By optimizing the communication, the overall efficiency and responsiveness of the system can be improved. Adaptive communication in the communication optimization strategy dynamically adjusts the communication frequency according to the network state and device requirements. In [43], Wang et al. proposed an adaptive communication mechanism that dynamically adjusts the communication frequency and reduces the communication overhead by monitoring the network and device states. Differential privacy in the communication optimization strategy protects data privacy by adding noise to the transmission parameters while reducing the amount of precise data to be transmitted. In [29], the DPSGD algorithm proposed by McMahan et al. protects user privacy by adding Gaussian noise. These techniques not only reduce the amount of communication but also increase the efficiency of communication at the same time. In [74], Guo et al. proposed a robust and efficient soft clustered federated system named REC-Fed, which aims to solve the problem of resource-constrained edge networks. The system enhances the personalization and robustness of model aggregation through a hierarchical aggregation method. In addition, adaptive model transmission optimization was also designed to jointly optimize model compression and bandwidth allocation to improve transmission efficiency. In [75], Ma et al proposed a novel FL framework. Joint optimization of computational and communication resources is achieved through computational offloading. The framework utilizes computational offloading to deal with the challenges posed by data heterogeneity. It also minimizes Kullback-Leibler (KL) scatter by optimizing computational offload scheduling. Minimizing communication costs through resource allocation. Federated learning based on computing Offloading decouples the optimization of computational and communication resource allocation into two steps. It effectively improved the convergence and accuracy of the model and reduced the negative impact of data heterogeneity on the system.

In FEL, model compression and communication optimization are interrelated. Model compression reduces the amount of data that needs to be transmitted, thus reducing communication overhead. Meanwhile, communication optimization can improve the communication efficiency and thus reduce the need for model compression. Therefore, in FEL, model compression and communication optimization are complementary and need to be considered and optimized comprehensively.

#### *4.1.3 Energy Consumption Sensing and Energy Optimization*

Energy-aware and energy-saving optimization is to reduce the energy consumption in the computation and communication process by means of an energy model and a low-power algorithm [43]. The energy-aware scheduling method dynamically adjusts the allocation order of computation tasks according to the real-time situation by means of the ED battery power and the computation load in order to balance the system devices, reduce the waiting time, reduce the energy consumption, and prolong the device's endurance [44]. The energy model in energy-aware scheduling is implemented by means of the establishing an energy consumption model to dynamically adjust the computation tasks according to the real-time energy consumption of the devices.

In [76], Li et al. proposed a task scheduling algorithm based on the energy consumption model to optimize the task scheduling strategy by monitoring the device power in real time. In addition, the energy-saving algorithm in energy-aware scheduling uses a low-power algorithm to reduce the energy consumption in the computation process. In [77], Wen et al. proposed an energy-efficient FL algorithm to improve the device endurance by optimizing the energy consumption during computation. Inter-device Collaboration and Energy Sharing in Energy Sensing and Energy Saving Optimization Achieve efficient use of resources and energy saving optimization through collaborative computation and energy sharing between devices. In [78], Stergiou et al. proposed an innovative cloud-based architecture, InFeMo, focusing on optimizing the allocation of computational and communication resources. InFeMo achieves efficient computational resource utilization by combining FL scenarios and existing cloud models with the flexibility of choosing to train the model on either a local client or a cloud server. This strategy not only reduces the waiting time for user requests, but also optimizes resource efficiency through energy-efficient design, further enhancing the coordinated allocation of computing and communication resources.

Collaborative computing is inter-device collaborative computing, where a task or model training is accomplished together through the decomposition and sharing of tasks and communication collaboration between ED [79]. Inter-device collaboration can fully utilize the computational resources of ED to improve computational efficiency and system performance, while reducing the energy consumption of individual devices [80]. In [42], Liu et al. proposed a FL framework based on collaborative computing, which improves computational efficiency through inter-device task collaboration. Energy sharing is an energy saving optimization through energy sharing and dynamic scheduling among ED. Energy sharing can improve the overall energy efficiency of the system and extend the device endurance while reducing energy consumption. In [81], Zhang et al. proposed an energy sharing mechanism that optimizes computational resource utilization through energy scheduling between devices.

In FEL, energy sensing and energy saving optimization is a complex process that requires comprehensive consideration and optimization of factors such as energy sensing scheduling, inter-device collaboration and energy sharing. By reasonably assigning tasks and scheduling, the overall performance and efficiency of the system can be improved and the optimal utilization of resources can be achieved. At the same time, through inter-device collaboration and energy sharing, the energy consumption of individual devices can be reduced, the duration of the devices can be extended, and energy-saving optimization can be achieved.

### *4.2 Joint Data Selection and Communication Resource Allocation for FEL*

In FEL, due to the heterogeneity of ED and resource constraints, how to effectively select training data and optimize communication resources becomes the key to improve the performance of FEL. In this paper, we explore the strategy of joint data selection and communication resource optimization based on FEL and its application to achieve efficient resource allocation and system performance improvement.

In FEL, the data on ED are often non-independently and identically distributed (non-IID), and there are large differences in data distribution and data volume across devices. An effective data selection strategy not only reduces the communication overhead, but also ensures the global convergence and performance of the model. Therefore, reasonable data selection is crucial for the successful implementation of FEL. In addition, communication resources are one of the key factors affecting system performance. An efficient communication resource optimization strategy can reduce the transmission delay, lower the energy consumption, and improve the transmission efficiency, thus enhancing the overall performance of FEL.

The integration of joint data selection and communication resource optimization strategies aims to construct an efficient optimization framework to achieve the co-optimization of data selection and communication resource allocation. The joint data selection and communication optimization model is constructed to achieve overall optimization by comprehensively considering data characteristics and communication costs. By incorporating data selection and communication resource optimization into the same optimization model, the global optimization is achieved by comprehensively considering data distribution, computational load and communication cost. Current research on sensing technology mainly focuses on a single federated device, ignoring the competition between devices and the resource allocation problems within devices, which limit the application of sensing technology.

To address this challenge, in [3], Fu et al. first delved into the potential bottlenecks when executing multiple federated tasks and constructs a federated optimization model to model the problem as a multidimensional optimization problem, which involves the device selection and communication resource allocation in a two-stage Stackelberg game. In order to solve this problem more efficiently, we propose a device selection and resource allocation method based on a multi-coalition game, which ultimately proves that the proposed method can reduce the training time and save the communication cost. The traditional centralized learning approach transmits data directly to the center for processing, which, although simple, introduces significant communication delays and may raise the risk of serious privacy breaches. To overcome these challenges, a significance-aware FEL system has been proposed in [82]. The system aims to enhance learning efficiency by optimizing end-to-end latency. By analyzing the relationship between communication resource allocation and data selection, and by exploiting the correlation between loss attenuation and gradient paradigm, an optimization model aiming to maximize the learning efficiency is constructed. Based on this, a data selection strategy and a communication resource allocation method are further developed that achieves optimal performance for a given end-to-end delay and sample size. By using a golden section search algorithm with low computational complexity, the optimal end-to-end delay setting can be determined. Experimental results on three popular convolution neural network (CNN) models show that the scheme not only significantly reduces the training latency but also improves the learning accuracy compared to other benchmark algorithms.

In [83], Liu et al. developed a new user scheduling algorithm for data collection in EL called data importance-aware scheduling. A key feature of this scheduling algorithm is that the informativeness of the data samples is taken into account, in addition to communication reliability. Specifically, the scheduling decision is based on the data importance indicator (DII), which elegantly combines two "importance" metrics from communication and learning perspectives, namely, signal-to-noise ratio (SNR) and data uncertainty. The scheme can intelligently perform joint selection of channels and data for training data upload to accelerate learning. In FEL, due to frequent model updates, the system needs to adapt to the limited communication bandwidth, the limited energy source of the ED, and the statistical heterogeneity of the ED data distribution. Therefore, the subset of devices used for training and uploading models must be carefully scheduled. Compared to previous work on FEL, data properties have been under-explored, and for this reason, in [84], Taïk et al. proposed a new scheduling scheme for FEL with non-IID and unbalanced datasets. Because data is a key component of learning, a new set of factors are considered for taking data properties into account in the FEL wireless scheduling algorithm. In the proposed algorithm, both data and resource perspectives are considered. In addition to minimizing the completion time of FEL and the transmission energy of the participating devices, the algorithm prioritizes devices with rich and diverse data sets. Ultimately, this strategy helps to reduce cost and improve the efficiency and performance of FEL. Sometimes, limited wireless communication resources greatly restrict the number of participating users and are considered to be the main bottleneck hindering the development of FEL. To address this problem, in [85], Jiang et al. proposed a user selection strategy for FEL systems based on data importance. In order to quantify the data importance of each user, the relationship between loss attenuation and the squared gradient paradigm is first analyzed. Then, a combined optimization problem is formulated to maximize the learning efficiency by jointly considering user selection and communication resource allocation. Through problem transformation and relaxation, the optimal user selection strategy and resource allocation are obtained, and a polynomial time optimal algorithm is given. Finally, simulations are performed with DNN models. The experimental results show that the algorithm has strong generalization ability and can obtain higher learning efficiency compared with other traditional algorithms. In [86], Xu et al. proposed a multi-intelligence reinforcement learning algorithm for optimizing data selection and communication resource allocation in intelligent cyber-physical systems (ICPS). By modeling the resource allocation problem among heterogeneous devices as a Stackelberg game and utilizing a partially observable Markov decision-making process, the algorithm efficiently optimizes data selection and communication policies among participating devices without sharing private information. The method reduces the differences in policy evaluation caused by interactions between devices and significantly improves the convergence speed of the system, ensuring efficient allocation of data and communication resources.

In FEL, the unreliability of the wireless channel may lead to random errors in the packets, which in turn has a significant impact on the convergence speed and learning delay of the model. To address this challenge, in [87], Xu et al. proposed an adaptive modulation strategy that aims to balance the learning delay and convergence speed caused by random channel errors. Unlike the traditional fixed modulation, this new scheme allows the wireless FEL system to dynamically adjust the modulation based on the computational capability of the device, the channel condition, and the relative importance of the training data, i.e., to achieve the joint consideration of the data importance and the optimization of communication resources. To enhance the sensing performance, an optimization algorithm for joint spectrum allocation and modulation scheme selection is further proposed, aiming to maximize the learning efficiency. Experimental results confirm that the proposed FEL framework can significantly improve the convergence speed of model training and thus the overall learning efficiency. For NR-U based industrial IoT networks under, in [88], Chen et al. proposed an innovative communication efficient FEL mechanism. The mechanism aims to select industrial IoT devices with high importance to data for local training under some relatively abundant resources. The objective function aims to balance the relationship between the importance of total data and transmission delay in FEL, which is achieved through joint learning, device selection and resource management scheduling. The Gradient norm value (GNV) of the local model of industrial IoT devices is used as the data importance indicator. When dealing with the Mixed Integer Nonlinear Programming (MINLP) problem, the Alternating Direction Method of Multipliers with Block Coordinate Update (ADMM-BCU), which has a low computational complexity, is used. ADMM-BCU algorithm, which is capable of deriving closed form expressions for optimal device selection and resource management. The final results show that the proposed strategy is able to accelerate the training process, improve the accuracy of the FEL, and significantly enhance the system efficiency. In [89], Albaseer et al. introduced new algorithms for running semi-supervised FL at the edge of the network, where devices have scarce labeled data and abundant unlabeled data. Considering the limited computational and communication resources, as well as the deadline constraints specified by the system, a dichotomous based algorithm is finally proposed to minimize the energy consumption and find the optimal transmit power and local CPU speed. Three control algorithms are then proposed to automatically label the unlabeled data samples during the training round. All algorithms use strong data augmentation during the training phase and weak data augmentation during the pseudo-labeling phase. Ultimately the algorithms effectively utilize the unlabeled data samples. In [9], He et al. Due to limited communication resources, traditional centralized learning methods by directly transmitting data can lead to significant communication delays and may pose a serious risk of privacy leakage. To address these challenges, the FEL framework is explored and a novel joint selection and resource allocation strategy that takes into account the importance of data is designed with the aim of improving the efficiency of the learning process. By integrating the allocation of communication resources and the selection of data, an optimization scheme with low computational complexity is proposed. Experimental results show that the scheme can significantly reduce the delay in the training process and improve the accuracy of the model.

In the hierarchical FEL architecture, in [8], Qiang et al. proposed a hierarchical FEL strategy with data importance awareness. This strategy aims to maximize the learning efficiency of hierarchical FEL by optimizing data selection and resource allocation. To achieve this goal, a joint algorithm that integrates the importance of data is designed. To solve the problem of data selection and resource allocation, it is decomposed into three sub-problems: ED association, resource allocation and data selection. For each sub-problem, a corresponding processing method is made. By correctly selecting important data, optimizing resource allocation and reasonably associating ED, the convergence speed can be significantly accelerated, which in turn significantly improves the learning performance. The selection of Hierarchical Federated Learning (HFL) nodes affects the quality of model training. In [56], Xin et al. investigated the optimization problem of node selection accuracy in HFL. To improve the quality of model training, a reputation-based node selection algorithm is designed. In the reputation-based node selection algorithm, ES selects nodes with high reputation prediction value to participate in model training, and nodes select neighboring nodes with high transmission capacity to collaborate. In [90], Chen et al. proposed a new Semi-Asynchronous Hierarchical Federated Learning (SAHFL) framework for mobile edge networks to enable resilient edge cloud model aggregation from data sensing. We further formulate a federated edge node association and resource allocation problem proposed under the SAHFL framework to prevent the individuality of heterogeneous devices and achieve communication efficiency. The scheme is finally demonstrated to speed up the training process and improve the performance of mobile edge networks.

Intelligent systems across application domains face challenges such as data heterogeneity, limited wireless resources, and device heterogeneity, which require intelligent participant selection schemes to speed up convergence. In [91], Albaseer et al. propose joint participant selection and bandwidth allocation schemes to address these challenges. A relaxation method is utilized to handle the combined nature of participant selection, making the complex constraints less stringent. Subsequently, a prioritized selection algorithm is developed to select optimal participants with low time complexity using relaxation-based solutions. The results show that the solution improves data utilization and speeds up convergence. In [92], Albaseer et al. go on to propose a new cluster FEL client selection method that exploits the heterogeneity of the devices to schedule clients based on their round delays and utilizes bandwidth reuse for clients that take more time to update their models. The server then performs model averaging and clusters the clients based on predefined thresholds as a way to reduce training time and speed up convergence. In FEL, energy-constrained devices at the edge of the network consume a large amount of energy when training and uploading local ML models, leading to a shortened lifetime. In [93], Albaseer et al. consider how to find the optimal user's resources, including fine-grained selection of relevant training samples, bandwidth, transmission power, beam forming weights, and processing speed objectives, to minimize the total energy consumption during communication rounds in FEL for a given deadline constraint. Since data heterogeneity degrades the performance of FL and reduces the resource utilization, then in [51], Serhani et al. proposed a heuristic Dynamic Edge Selective Scheduling algorithm (DSS-Edge-FL) aiming to optimize the resources and address the data heterogeneity. The final experimental results show that the method can speed up convergence and improve resource efficiency, since the server needs to first capture all data distributions from all clients to perform correct clustering. Due to resource and time constraints at the edge of the network, only a small number of devices are selected in each round, so efficient scheduling techniques are needed to address these issues, so in [94], Albaseer et al. newly proposed new client scheduling and selection algorithms for cluster federated multitasking learning in order to improve the convergence speed while capturing all the data distributions, thus improving the model for each cluster. This approach ensures proper clustering and fairness among clients by utilizing bandwidth reuse, and the results show that the proposed algorithm reduces the training time and improves the convergence speed significantly. In [95], Hu et al. proposed a clustered data sharing framework to optimize the processing of non-IID data. Through side-link-assisted multicasting, the framework selectively shares data to mitigate the data heterogeneity problem. In addition, model convergence is accelerated by an adaptive clustering algorithm and a stochastic optimized resource allocation algorithm. The model accuracy in a limited communication resource environment is improved. The application of joint data selection and communication resource allocation is effectively demonstrated.

In the FEL system, vehicles upload data to the ES, which trains the vehicles' data to update the local model [96] and then returns the results to the vehicles to avoid sharing the raw data [97,98]. However, the cache queue at the edge is limited and the channel between the ES and each vehicle is time-varying [99]. Therefore, it is a challenge to choose the right number of vehicles to ensure that the uploaded data maintains a stable cache queue in the ES [100] while maximizing the learning accuracy. In addition, selecting vehicles with different resource states to update the data affects the total amount of data involved in training, which in turn affects the accuracy of the model [101,102]. Therefore, in [103], Wu et al. proposed a vehicle selection scheme for FEL systems in in-vehicle networks. This scheme considers the states of all vehicles in the coverage area to maximize learning accuracy while ensuring cache queue stability amid the influx of diverse data in 6G environments. In [104], Zhou et al. proposed an FL model for efficient learning in a distributed end-edge-cloud setup, focusing on data privacy and lowering communication costs. The two-tier design maximizes data use from vehicles and RSUs. Table 1 summarizes the details of the review in this subsection and shows the highlights of this section.

**Table 1:** Summary of reviewed works

| Literature | Algorithm deployment location | Advantages and disadvantages | Main technologies | Impact of resource optimization on other system performance |
|---|---|---|---|---|
| [3], [82], [83], [84], [85], [86] | ED and cloud | Data selection and resource allocation strategies perform well in edge environments but may be limited in high-complexity tasks. | Focuses on optimization of ED such as data selection and user scheduling | Optimization improves edge computing response time and computational efficiency but may negatively impact data transmission and system stability. |
| [87], [88], [89], [9] | ED and IoT networks | High-complexity algorithms may incur higher computational and communication costs but offer significant performance optimization advantages. | Includes advanced technologies such as adaptive modulation and semi-supervised learning | Enhances communication efficiency and model accuracy but may increase computational load when handling large-scale data. |
| [8], [56], [90] | Hierarchical edge computing environments | Hierarchical learning methods effectively optimize resource allocation but may introduce additional computational complexity and synchronization issues. | Includes hierarchical learning and reputation-based node selection techniques | Improves efficiency in hierarchical learning but may affect system synchronization and stability. |
| [91], [92], [93], [51], [94], [95] | Heterogeneous edge computing networks and clustered environments | Different approaches have varied advantages and disadvantages regarding energy efficiency, response time, and fairness, with significant optimization effects in these areas. | Data-driven selection, fine-grained selection, and dynamic scheduling technologies | Improves system energy efficiency and response time but may impact system fairness. |
| [96], [97], [98], [103], [104] | Vehicular networks and heterogeneous edge computing environments | Applications in vehicular networks and heterogeneous environments may face stability and accuracy issues, but overall optimization effects are notable. | Includes mobility-aware caching, asynchronous learning, and multi-agent reinforcement learning | Enhances stability and information timeliness in vehicular networks but may require additional computational resources in dynamic environments. |

The efficient allocation of joint data selection and communication resources based on FEL, with a wide application coverage, can significantly improve the overall system performance and resource utilization efficiency through the integration of intelligent data selection and communication optimization strategies. Future research can conduct in-depth exploration in terms of smarter resource optimization algorithms, real-time dynamic adjustment strategies, and multi-level collaboration mechanisms to further enhance the performance and scalability of the FEL system.

### *4.3 Joint Scheduling and Communication Resource Allocation for FEL*

FEL plays an important role in distributed ML, and by performing local training on ED and transmitting only model parameters instead of raw data, FEL effectively protects data privacy. However, the heterogeneous and resource-constrained nature of ED, as well as network bandwidth limitations, make how to efficiently perform task scheduling and communication resource allocation a key issue in improving the performance of FEL. In this chapter, we will explore the joint scheduling and communication resource allocation strategy based on FEL and its applications, aiming to construct an efficient optimization framework to achieve collaborative performance optimization of scheduling and communication resource allocation.

In FEL, model convergence speed can be improved by optimizing scheduling and resource allocation strategies to reduce training time. Due to the heterogeneity of the training data distribution, enhancing the convergence speed can improve the training efficiency and performance of the model. In [105], Shi et al. formulate the bandwidth allocation and scheduling problem to improve FL convergence. They decouple it into two sub-problems: for bandwidth allocation, they find that more bandwidth should be given to devices with poorer channels or weaker computational abilities. For scheduling, a greedy strategy is used to balance training rounds and latency, selecting devices with the fastest model updates to optimize learning efficiency. In [106], Yin et al. address convergence delay with a client scheduling and resource allocation algorithm that considers system and client heterogeneity, using an adaptive bandwidth allocation strategy. In [107], Shi et al. propose a strategy that combines device scheduling and resource allocation to maximize FL model accuracy within a fixed training time, balancing per-round delay and the number of training rounds. In [108], Ren et al. focus on FEL gradient averaging, proposing a probabilistic scheduling strategy that considers channel quality and the importance of updates. This framework achieves faster model convergence and higher accuracy compared to traditional methods.

Airborne Computing (AirComp) shows potential as a future solution by exploiting the superposition properties of wireless channels. However, fading and noise in wireless channels may lead to distortion in aggregation during FL, thus affecting the accuracy of the model. In addition, the quality of the data and the energy consumption of the ED can also affect the efficiency of model aggregation and convergence. To address these issues, in [109], Du et al. proposed a dynamic device scheduling mechanism that is able to select eligible ED with appropriate power control strategies for transmitting their local models to participate in server-side model training in FL. In this mechanism, the importance of the data is jointly measured by the gradient of the local model parameters, the channel conditions and the energy consumption of the device. In particular, in order to fully utilize the distributed datasets and speed up the convergence of FL, the mechanism retains and accumulates the local updates from the distributed devices for potential future transmissions rather than discarding them directly. In addition, a strategy for searching the best device selection is developed through a Liapunov drift plus penalty optimization problem. Simulation results show that the scheduling mechanism has higher test accuracy and faster convergence speed, and is robust to different channel conditions. This suggests that the mechanism is able to effectively deal with various challenges that may arise in FL, thus improving the accuracy and learning efficiency of the model. Since the wireless devices involved in FEL have limited resources in terms of communication bandwidth, computational power and battery capacity, their scheduling must be carefully designed to optimize the training performance. In this study, in [110], Sun et al. highlight the importance of over-the-air FEL systems with simulated gradient aggregation and propose an energy-aware dynamic device scheduling algorithm to optimize the training performance of the devices within energy constraints.

In FEL, enhancing the FEL performance can improve the generalization ability and robustness of the model by optimizing the scheduling and resource allocation strategies, thus improving the performance and effectiveness of the model. Due to frequent communication, FEL needs to adapt to the limited communication bandwidth. In addition, the statistical heterogeneity of local datasets distributions and the uncertainty of data quality pose important challenges to the convergence of training. Therefore, careful selection of participating devices and similar bandwidth allocation are necessary. In [111], Taïk et al. proposed a data quality-based scheduling algorithm that prioritizes reliable devices that have rich and diverse datasets. In this way, the algorithm aims to improve the performance of the FEL system and to ensure that the FL process can be carried out efficiently in resource-constrained environments. Furthermore, in [112], Wen et al. designed a training algorithm for a hierarchical FEL system, which consists of phases such as local gradient computation, weighted gradient uploading and ML model updating. By mathematically describing these phases and analyzing the convergence bounds of a single round of the training algorithm, a design problem involving scheduling and resource allocation schemes is proposed. The design problem aims to simultaneously take into account the uncertainty of the wireless channel and the importance of the weighted gradient, so as to effectively mitigate the privacy risk and communication overhead, and improve the overall performance of the FEL system. In [113], Wen et al. mathematically modeled device availability, wireless channel quality, and gradient quality and derived convergence bounds for model training in FEL systems, and based on the analysis, formulated a joint device scheduling and resource allocation problem aimed at improving FEL efficiency and performance.

In FEL, due to the energy constraints of ED, minimizing energy consumption can extend the device range, reduce energy consumption, and improve system efficiency. In [114], Hu et al. examined the use of FEL in cellular networks and proposed a strategy to reduce energy consumption while maintaining model performance. This approach makes FEL more suitable for real-world applications, offering a flexible and adaptable solution for different scenarios. One of the key challenges addressed is balancing learning tasks with extending device battery life. In [34], Feng et al. explored combining Heterogeneous Computing (HC) and Wireless Power Transfer (WPT) in FL to improve energy efficiency. They formulated an optimization problem to minimize smart device energy consumption while maximizing energy acquisition and solving the heterogeneous scheduling problem. This included optimizing WPT timing, dataset size, transmission power, sub-carrier allocation, and processor frequency. The solution was derived by decoupling the variables for high efficiency. Simulations showed improved energy efficiency and fast convergence. In [115], Albaseer et al. proposed a scheduling strategy to balance energy consumption across devices, ensuring fairness and faster convergence. In [116], Ozfatura et al. explored the implementation of FEL in wireless fading channels, addressing challenges from downlink and uplink delays and random computational delays. They accelerated model training by overlapping communication and computation using fountain-encoded global model updates, allowing clients to asynchronously start local computations. A dynamic scheduling policy, Minimum Remaining Time-based Policy (MRTP), was proposed for uploading local updates, focusing on minimizing upload time. To address biases in non-IID data scenarios, they introduced two fairness-focused alternatives: age-aware MRTP (A-MRTP) and opportunity-fair MRTP (OF-MRTP). OF-MRTP significantly reduced delays while maintaining high accuracy, as confirmed by simulations. In [117], Hu et al. addressed challenges from the time-varying nature of wireless channels that affect training delay and energy consumption. They developed a dynamic scheduling and resource allocation algorithm for streaming data scenarios, where new data samples are generated over time. Using a stochastic network optimization approach with a Lyapunov drift-plus-penalty framework, the algorithm adapts device scheduling, computational capacity, bandwidth, and transmit power. Results demonstrated enhanced learning performance and improved energy efficiency. In [118], Liu et al. introduced a hybrid split and FL framework for wireless UAV networks, addressing user resource diversity and computational capacity. Users can choose between split and joint training methods. To handle unreliable channels and limited energy, they modeled scheduling and method selection as a multi-choice knapsack problem. Their energy-efficient algorithm selects users each round, allowing method choice. Simulation results showed that this framework significantly reduces energy consumption while maintaining test accuracy, underscoring its efficiency in resource-constrained environments. Table 2 summarizes the details of the review in this subsection and shows the highlights of this section.

**Table 2:** Summary of reviewed works

| Literature | Algorithm deployment location | Advantages and disadvantages | Main technologies | Impact of resource optimization on other system performance |
|---|---|---|---|---|
| [105], [106], [107], [108] | Wireless federated edge networks | Strategies offer fast convergence and manage heterogeneity effectively, but may struggle with non-IID data and latency constraints. | Focuses on client scheduling, wireless resource allocation, and channel awareness. | Improves system latency and convergence speed but may impact overall communication efficiency and fairness in heterogeneous environments. |
| [109], [110] | Federated edge learning systems with over-the-air computation | Dynamic scheduling optimizes gradient and channel awareness but can be complex and energy-intensive. | Incorporates gradient and channel awareness, energy constraints. | Enhances scheduling efficiency and energy use but may increase computational and scheduling complexity. |
| [111], [112], [113] | Hierarchical and quality-based federated edge environments | Quality-based scheduling improves energy efficiency but may introduce complexity in managing hierarchical systems. | Emphasizes data quality, availability, and hierarchical scheduling. | Improves energy efficiency and scheduling quality but may affect system's overall hierarchical performance. |
| [114], [34], [115], [116], [117], [118] | Energy-efficient federated edge networks and hybrid environments | Focuses on balancing energy consumption and optimizing scheduling, but may struggle with dynamic environments and streaming data. | Includes energy-efficient scheduling, historical participation, and Lyapunov optimization. | Enhances energy efficiency and scheduling but may introduce complexity and affect system stability in dynamic environments. |

Joint Scheduling and Communication Resource Allocation Based on FEL summarizes the importance of joint scheduling and communication resource allocation in FEL through the integration of dynamic task allocation, priority scheduling, model compression and quantization, dynamic bandwidth allocation and so on, as well as how to improve the convergence speed, minimize the energy consumption to reduce the communication overhead and enhance the performance of FEL through the optimization of the scheduling and resource allocation strategies. Future research can explore in-depth on smarter resource optimization algorithms, real-time dynamic adjustment strategies and multi-level collaboration mechanisms to further enhance the performance and scalability of FEL systems.

### *4.4 Joint Network Topology and Communication Resource Allocation for FEL*

In FEL, network topology and communication resource allocation involve how to rationally design the network topology and how to effectively allocate communication resources to support model training and inference. Optimizing network topology and communication resource allocation is the key to improving the overall performance of FEL. In this section, various network topology and communication resource allocation issues in FEL are explored. Through the study, we hope to provide theoretical support and practical guidance for the deployment and development of FEL in real applications.

Network topology plays an important role in FEL as it determines how data is shared and transmitted between devices. Transmit power optimization is the process of adjusting the transmission strength of wireless signals to improve communication. In FEL, transmit power optimization helps to reduce communication delays and improve the reliability of data transmission. Transmit power optimization refers to adjusting the transmission strength of wireless signals to improve communication. In FEL, transmit power optimization helps to reduce communication delay and improve the reliability of data transmission. The communication frequency is dynamically adjusted according to the real-time status of the device and the network to optimize the communication resource utilization. In [119], Lim et al. proposed a reinforcement learning based communication frequency optimization method to achieve adaptive optimization of communication resources by dynamically adjusting the communication frequency. The joint optimization of the adaptive network topology and the communication transmission frequency can optimize the resource utilization more. In [120], Zhang et al. proposed an innovative device selection method for the impact of inter-cell interference on the performance of AirComp based FL in large-scale cellular networks. The method first delineates an interference reduction region within each cell and selects the devices involved in FL from it. At the same time, the devices that may cause interference in each cell are set to be silent on the resource block used by AirComp as a way to mitigate the interference between cells. In addition, the effects of path loss and small-scale fading will be overcome by adjusting the transmit power. In order to fully evaluate the performance of the proposed scheme, different network topology of base stations and devices with different spatial distribution characteristics are also considered and valid and realistic datasets are used in the neural network-based FL experiments. The experimental results show that this approach not only significantly improves the average prediction accuracy of FL, but also effectively reduces the inter-cell interference.

Task offloading techniques optimize the utilization of computational and network resources and improve system performance by offloading computational tasks from the ED to the ES or cloud [121]. The computation offloading policy determines which tasks should be offloaded to the ES for execution to optimize system performance. In [122], He et al. explored how compute offloading techniques in MEC can effectively deal with problems such as core network congestion and mobile device resource constraints. However, these solutions often fail to adequately consider user mobility and the uncertainty of the MEC environment. To this end, an innovative architecture is proposed for the first time that combines digital twin (DT) technology with MEC and FL frameworks. DT networks can virtually simulate the states of physical entities and network topology, enabling real-time data analysis and network resource optimization. Meanwhile, the computational offloading technique of MEC is utilized to alleviate the resource constraints of mobile devices and the congestion problem of the core network. FL is further utilized to construct a DT model based on the operational data of physical entities and network topology, which is jointly optimized for the computation offloading and resource allocation problems, aiming to reduce the "laggard" effect in FL. Ultimately, this scheme significantly reduces the total cost by about 50% and improves the communication performance. In [123], Huang et al. proposed an HFL framework designed for the Space-Air-Ground Integrated Network (SAGIN). The framework utilizes aerial platforms and Low Earth Orbit (LEO) satellites as multi-layered ES and CS to optimize the joint network topology through inter-satellite links. DRL and hybrid control strategies are applied in the study to equitably allocate resources and optimize aggregation weights in HFL. The method performs superiorly in resource allocation and network topology management, addressing the complex challenges of implementing HFL in SAGIN.

In a distributed FL environment without servers, the convergence performance of the algorithms is not only affected by the quality of communication and the number of edge nodes, but is also closely related to the graph topology of the communication network between these nodes. In real-world application scenarios, the convergence of distributed FL may also be constrained by quantization errors due to limited communication resources, e.g., the presence of latency and energy consumption.

In [124], Yan et al. studied the decentralized gradient descent (DGD) algorithm in distributed FL. They analyzed how graph topology and quantization influence DGD convergence under different wireless conditions and communication limits. The research identified the maximum quantization error that can ensure convergence and compared convergence bounds for high-connectivity versus low-connectivity topologies, revealing their differing energy consumption. The results emphasized the important role of graph topology in convergence under energy constraints. In [125], Huang et al. proposed a topology optimization scheme for FEL to address data heterogeneity and enhance efficiency. The method co-optimizes node aggregation topology and computational speed to minimize energy consumption and communication delay. They used an iterative convex approximation method for stable solutions and introduced an imitation learning approach with a DNN for real-time decisions. Simulation results demonstrated that the scheme accelerates FL training and significantly improves energy efficiency. In [126], Sun et al. proposed a semi-decentralized FEL framework that enhances training speed by leveraging diverse data from multiple edge clusters. Their training algorithm consists of local model updates and inter-cluster model aggregation. The study shows how network topology and communication overhead impact performance, with results indicating that this approach converges faster than traditional FL methods. In [17], Sun et al. introduce a semi-decentralized FEL framework for 6G networks that addresses non-IID data and resource allocation. The framework integrates local training and model aggregation across clusters to improve collaboration. To minimize model bias, multiple exchanges occur during aggregation. The study analyzes how model aggregation cycles and network topology affect convergence speed. Results indicate that sparse connections can slow convergence, but increasing model sharing can boost efficiency. In [127], Kavalionak et al. relied on centralized servers to control and manage the training process of ML models is not always feasible. Therefore, turning to explore the problem of training ML models on a network of nodes in a fully decentralized manner, the experiments used different network topology, datasets, and ML models to show how the tuning of these variables can speed up convergence. In [128], Chen et al. explore geographically distributed FL for future 6G networks, focusing on fully decentralized model training. This approach eliminates the need for a centralized server but may increase communication costs. To address this, they propose a synchronization interval optimization strategy for latency-constrained environments, aiming to maximize model accuracy within a set time frame and with limited resources. Their analysis derives convergence bounds and shows the algorithm's adaptability to data heterogeneity and network topology, enhancing convergence speed. Table 3 summarizes the details of the review in this subsection and shows the highlights of this section.

**Table 3:** Summary of reviewed works

| literature | Algorithm deployment location | Advantages and disadvantages | Main technologies | Impact of resource optimization on other system performance |
|---|---|---|---|---|
| [119], [120] | Mobile edge networks | Provides broad insights, lacks deep optimization for specific cases. | Inter-Cell Interference Coordination, AirComp | Improves communication efficiency, may increase inter-cell interference. |
| [122], [123] | DT-MEC assisted framework | Efficient integration with digital twins, complexity is a challenge. | Computation Offloading, Resource Allocation | Reduces latency, may increase computational and resource management complexity. |
| [124], [125], [126], [17], [127], [128] | Decentralized and Geo-Decentralized edge networks | Flexible, robust in static environments, synchronization issues in dynamic contexts. | Quantization, Graphical Topology, Synchronization Interval Optimization | Reduces communication overhead, affects synchronization and update frequency. |

The study of joint network topology and communication resource allocation for FEL focuses on designing an efficient network structure and reasonably allocating communication resources to optimize the performance of distributed model training, ensuring efficient data transmission and fast model convergence under limited bandwidth, delay and energy consumption, while taking into account privacy preservation and system security, by comprehensively considering the network conditions, device capabilities and data characteristics, with the aim of improve the communication efficiency, robustness and reliability of the whole FEL system.

## 5 Future Challenges and Perspectives

With the popularity of IoT devices and the explosive growth of data volume, the traditional centralized data processing model can no longer meet the demands of real-time and privacy protection. Therefore, FEL has been proposed to address these issues. FEL allows model training and inference to be performed on distributed devices at the edge of the network, which reduces data transmission latency, improves computational efficiency, and enhances privacy protection. However, FEL faces many challenges in its implementation, one of which is how to efficiently allocate and manage federated resources to ensure model training efficiency and performance. Although the optimization of multi-resource allocation strategies is the main discussed issue in our review, other equally important challenges do exist involving how to cope with data heterogeneity and imbalance, guarding against malicious attacks, device failures and network instability, referred to as model accuracy, security and robustness. These factors are equally critical in distributed machine learning environments and need to be further investigated and addressed in the future. Moreover, the introduction of an interdisciplinary approach can further optimize the performance of the FEL system in terms of joint resource allocation.

### *5.1 Joint Computing and Communications Resource Allocation*

In FEL, the computational load and communication requirements of ED may fluctuate rapidly due to the number of devices, types of tasks, and changes in the environment. In a dynamic network environment, devices may leave the network due to movement or other reasons. At this point, the system needs to dynamically adjust the allocation of computational and communication resources to ensure that other devices can continue to effectively participate in the FL process. However, traditional static resource allocation methods have been difficult to cope with such dynamics, so how to achieve dynamic load balancing and resource optimization has become an important challenge. In addition, ED have uneven computing power and storage resources, and it is also a challenge to efficiently allocate resources under such heterogeneous conditions. Future research can focus on developing intelligent resource scheduling algorithms, such as real-time monitoring and dynamic adjustment of ED based on DRL and distributed FL optimization methods [129]. These algorithms are capable of adaptive allocating computing and communication resources based on the current load state, energy efficiency requirements [130], and task priorities of the devices in order to optimize the overall performance and energy efficiency of the system, but also with privacy and security in mind.

In addition, the data transmission efficiency directly affects the speed and effectiveness of model training [131]. However, ED is usually limited by bandwidth and transmission speed, especially in mobile environments. It remains a challenge to realize more low-latency and high-efficiency communication while ensuring data security and privacy [132]. With the diversification of application scenarios and the increasing explosion of data volume, traditional data compression and transmission protocols may no longer be able to meet the demand for real-time and efficiency, which remains a great challenge for the future research. Future research can explore the application of more advanced communication technologies, such as 5G or 6G networks, to enhance the communication speed and stability between ED [133]. In addition, combining techniques such as differential privacy and secure multi-party computation, more secure and efficient data transmission mechanisms can be designed to safeguard data privacy while enhancing communication efficiency. The application of using Internet of Vehicles (IoV) technology can also be increased to make resources allocation in the dynamic networks more stable [134].

ED management and optimization is the basis for joint computing and communication resource allocation. How to intelligently manage and optimize large-scale ED, including device health monitoring, hardware and software upgrade management, and resource utilization maximization, is a complex and challenging problem. Future research can combine IoT technology or IoV technology and AI algorithms to design intelligent ED management systems [135]. These systems can monitor the status and operation of the equipment in real time, predict the load and energy consumption of the equipment, and then realize the optimized management of the ED through automation and intelligence to improve the stability and efficiency of the overall system [136].

In the context of globalization, it is a new challenge to design resource allocation strategies for cross-region and cross-domain collaboration to achieve joint computing and communication resource optimization on a global scale. Network conditions and policy restrictions in different regions may lead to inefficient resource allocation or data privacy issues. Future research could focus on the development of globalization-oriented joint optimization algorithms and protocols that take into account the special needs and constraints of different geographic locations and network environments. By developing flexible resource allocation strategies and security protocols, joint model training and data sharing on a global scale can be realized to promote the application and development of FEL technology on a global scale.

Interdisciplinary approaches can further optimize the performance of FEL systems. The combination of reinforcement learning and optimization techniques is a potential avenue. In the work[137], it combines DRL with service caching, communication and computational resource optimization to provide a flexible resource management framework. Specifically, DRL can be used to dynamically adjust resource allocation. In such a way, the proposed method in [137], it can cope with changes in device heterogeneity, network dynamics, and mission requirements, thereby improving the overall performance and system robustness of FEL.

### *5.2 Joint Data Selection and Communications Resource Allocation*

In FEL, participants are usually data owners distributed on different ED who may have sensitive personal or corporate data. How to effectively perform data selection and model aggregation while protecting data privacy remains an important challenge. New privacy-preserving techniques and security protocols need to be developed to address potential data leakage and model tampering risks. Traditional data encryption and privacy protection techniques may not be able to fully meet the needs of FL, especially the possible security threats to the data during the communication process. Future research could focus on developing more efficient and secure FL frameworks, such as differential privacy, further optimization of isomorphic encryption techniques, and the application of multi-party computation. These techniques can protect data privacy while allowing efficient data selection and model updating among ED, thus ensuring the security and trustworthiness of the whole system. As users become more aware of data privacy, future privacy protection techniques are likely to be more personalized and fine-grained. Users will be able to adjust the level of privacy protection according to their privacy preferences, such as selecting which data can be used for model training and the amount of data that can be shared, thus realizing the goal of both protecting privacy and promoting data sharing and intelligent analysis.

In practice, the data distribution on ED usually changes dynamically, and there may be significant imbalance in the data volume and data characteristics of different devices. It is a challenge to effectively perform data selection and communication resource allocation under such dynamic and uneven data distribution. Traditional data distribution strategies may not be able to fully utilize the data resources of all participants, resulting in slower model convergence or poor model accuracy. Different devices may collect different types and qualities of data, leading to uneven data distribution. How to effectively utilize these heterogeneous data and avoid model bias towards some specific data distributions is a key challenge. Future research could explore data selection strategies based on dynamic learning and adaptive algorithms, such as adaptive aggregation methods and dynamic weight adjustment techniques in FL. These methods can automatically and dynamically adjust the strategies for data selection and model updating based on the current data distribution and load of the ED, thus improving the overall performance and efficiency of the system. New algorithms can also be designed, e.g., through techniques such as re-weighting or oversampling, and these are able to sense and understand the imbalance of data distribution and adjust the model training process accordingly. In the future, migration learning techniques can also be utilized to solve the data imbalance problem by migrating a model that has been trained on datasets to another device with a different data distribution. New communication protocols are then researched and developed to reduce the communication for model updates, e.g., through compressed communication protocols or selective update protocols, etc.

ED management and optimization is critical for joint data selection and communication resource allocation. How to intelligently manage and optimize large-scale ED, including device health monitoring, hardware and software upgrade management, and resource utilization maximization, is a complex and challenging problem, and privacy protection and data security must be ensured to avoid leaking sensitive information. Future research could combine IoT technologies and automated management systems to develop automation tools and frameworks for FL system deployment, monitoring and maintenance, and to design intelligent ED management platforms. It can monitor the status and operation of the equipment in real time, predict the load and energy consumption of the equipment, optimize the management of the ED through automation and intelligence, and improve the stability and efficiency of the overall system. A more realistic and comprehensive simulation environment can also be established to evaluate and optimize end-to-end performance and facilitate the practical application of research results. Algorithms that can automatically adjust strategies according to changes in the real-time environment are then developed, such as automatically adjusting the frequency of model updates and selecting the most appropriate communication paths. Model pruning, quantization, and other light weighting techniques are used to reduce the size of the model and improve operational efficiency on resource-constrained devices.

ED is usually limited by issues such as energy supply and thermal management, and it is an important technical challenge to optimize energy consumption while ensuring performance, and to improve the energy efficiency and environmental sustainability of devices. Joint data selection and communication resource allocation should take energy consumption and environmental impact into account to realize green intelligent edge computing. That is to say how to ensure the computational performance while reducing the environmental impact, such as reducing carbon emission and e-waste generation. Future research can start from both hardware optimization and algorithm innovation to develop low-power and high-efficiency ED, as well as researching and developing new energy supply strategies, such as using renewable energy sources (e.g., solar energy, wind energy) to power ED. In addition, comprehensive environmental sustainability can be achieved by considering the impact of energy consumption and the environment throughout the entire life cycle of the device, from design, manufacturing, use to recycling. Data selection and communication scheduling algorithms for energy consumption awareness are also designed. Intelligent energy management of the ED is realized by using energy prediction and dynamic power management techniques to enhance the energy efficiency and sustainability of the system.

Potential interdisciplinary approaches can be drawn from the combination of machine learning and communication theory. In [138], Jia et al. proposed low-complexity suboptimal algorithms. Optimizing resource allocation and data selection using matching theory and gradient projection methods. This approach not only improves system efficiency, but also reduces transmission and computation costs. In the future, the uncertainty and dynamic changes in resource allocation can be handled more effectively by using dynamic system control theory and game theory methods.

### *5.3 Joint Equipment Scheduling and Communications Resource Allocation*

As the scale of FEL technology expands and application scenarios diversify, the scalability and flexibility of the system becomes a key challenge. On the one hand, as more and more ED and service providers join the FL network, the system needs to be able to handle large-scale devices and services, which requires the system to be highly scalability. On the other hand, due to the diversity of application scenarios, the system needs to be able to adapt to different application requirements and environments, which requires a high degree of flexibility. How to design efficient system architecture and interface standards to support the integration of different vendors and service providers while ensuring system stability and performance optimization is a direction that requires in-depth research. Future research can facilitate multi-party participation and resource sharing by developing an open FL platform and modularization system architecture. Cloud-native technologies and micro-service architectures are introduced to build a flexible edge computing platform that can dynamically adjust and extend system functions, improve system adaptability and manageability, and support joint device scheduling and communication resource allocation in complex scenarios.

FEL involves collaboration and cooperation among multiple organizations or enterprises, each with different datasets and ED. how to achieve effective device scheduling and resource allocation across organizations to ensure balanced benefits and efficient cooperation among all parties is a key challenge. This collaboration and cooperation involved not only optimization at the technical level, but also coordination and cooperation at the management level. In addition, managing resource sharing and data collaboration among different organizations may be affected by a variety of factors such as laws and regulations, business competition, and privacy protection, increasing the complexity of management and coordination. Future research can start from the perspective of organizational management and collaboration mechanisms to design adaptable, open and transparent FL platforms. In order to achieve fair cooperation and resource sharing among cross-organizations, the trust framework of blockchain can be introduced to establish a trusted data exchange and resource allocation mechanism to promote fair cooperation and resource sharing among cross-organizations. Blockchain technology has the characteristics of decentralization, high transparency and non-tampering, which can effectively guarantee the security and fairness of data exchange. Through blockchain technology, the full tracking and verification of data in the FL process can be realized, ensuring that the source and destination of data can be traced and preventing data from being tampered with or leaked. In addition, smart contracts and multi-party secure computing technologies are developed to ensure data privacy and security, and provide legal and technical guarantees for joint equipment scheduling and communication resource allocation. Smart contracts are automated execution protocols based on blockchain technology, which can automatically execute agreements between two parties, reduce human intervention, and improve cooperation efficiency. Multi-party secure computing technology, on the other hand, can realize data analysis and computation between multiple parties without compromising privacy and guarantee data security.

FEL involves a variety of devices, platforms and communication protocols, and system integration and application deployment may face difficulties due to the lack of harmonization of technical standards and interoperability. This lack of harmonization may result in the inability of smooth communication and collaboration between devices, which in turn affects the performance and efficiency of the entire system. Therefore, the development and promotion of uniform technical standards to achieve connectivity between devices is an urgent issue for the successful implementation of FEL systems. Future research can advocate open and common FL standards to promote joint participation and contribution from all parties. This includes the development of unified data formats and communication protocols to support data exchange and task collaboration among different vendors and platforms. Encourage cooperation among different standardization organizations, such as the International Organization for Standardization (ISO), the International Electrician Commission (IEC), and the Open Mobile Alliance (OMA), to jointly promote the development and implementation of FL standards, and establish cross-disciplinary research teams, including computer scientists, communication engineers, data scientists, and legal experts, to work together to research and develop FL standards. Global technical frameworks and interoperability tests can also be established to ensure that FL systems between different vendors and platforms can collaborate smoothly, and then promote the standardization and application of joint device scheduling and communication resource allocation on a global scale.

FEL involves the processing and sharing of large amounts of sensitive data, and it is a critical challenge to build user trust and legal compliance while protecting user privacy. With increasingly stringent data protection regulations and users' heightened concerns about data security, the issue of legitimacy and transparency in the management and use of data across organizations becomes a future challenge. Future research can establish traceable and controllable data management mechanisms by enhancing compliance with data privacy protection technologies and privacy protection laws. For example, data privacy can be protected by adding noise to the datasets so that any analysis of the data cannot infer the private information of a specific individual, as well as multi-party secure computing frontiers, which are technologies that allow multiple parties to jointly analyze data without knowing the content of the other party's data, to ensure data security and privacy protection during transmission and processing. At the same time, user education and participation are being carried out to improve users' understanding of and trust in data use and sharing mechanisms, and to promote social acceptance of joint equipment scheduling and communications resource allocation.

In [139], interdisciplinary approaches can be applied in resource allocation for FEL, where it employed semantic communication in adaptive network management. Semantic communication technology helps devices understand the meaning of data transmission, thus reducing unnecessary communication burden. It improves the utilization of scheduling and communication resources and optimizes device scheduling and resource management in FEL.

### *5.4 Joint Network Topology and Communications Resource Allocation*

FEL involves different types and geographically distributed ED and network nodes. It is a key challenge to achieve effective network topology design and communication resource allocation in heterogeneous network environments to support data sharing and task collaboration across organizations. Heterogeneous networks may cover a variety of communication technologies such as Wi-Fi, 5G, LPWAN, etc., and how to achieve seamless integration and optimal configuration of these networks to improve the overall performance and responsiveness of the system is a problem that needs to be addressed. Future research can focus on developing adaptive network management and intelligent optimization algorithms to address the challenges in heterogeneous network environments. Real-time monitoring and optimization of edge network topology and communication resource allocation can be achieved by introducing ML and data-driven approaches. For example, DRL algorithms are utilized to dynamically adjust network topology and resource allocation strategies to maximize system performance and efficiency [140,141]. The future network topology design can use hybrid communication technology to dynamically allocate communication resources according to different application scenarios and network loads. For example, 5G technology is used in areas with high-density data transmission, while LPWAN technology is used in scenarios with high demand for low-power long-distance communication. This dynamic provisioning can effectively balance the bandwidth, latency and energy efficiency of the network, and improve the overall system performance and user experience.

With the popularity of edge computing and the diversification of application scenarios, future research will pay more attention to the virtualization technology and resource elastic allocation of edge networks. By introducing network function virtualization (NFV) and software-defined network (SDN) into the edge environment, elastic allocation and flexible deployment of network resources can be achieved. This virtualization technology can dynamically configure network services according to application requirements, improve network scalability and flexibility, and at the same time reduce network operating costs and energy consumption, providing technical support for the rapid development of future edge networks.

Future developments should also focus on the security and privacy protection of edge network communications. More advanced encryption technologies, authentication mechanisms, and security auditing tools are introduced to cope with the increasingly sophisticated threats of network attack and data leakage. With enhanced security measures, user trust in data sharing and edge computing services can be boosted, promoting the widespread adoption and sustainable development of FEL technology.

In [142], the dynamic beam control and resource allocation method proposed by yuan et al. can provide interdisciplinary application ideas for FEL systems, where combining graph neural networks with reinforcement learning, to adaptively deal with the network topology changes and resource allocation in FEL. Improving system adaptability and performance.

## 6 Conclusion

FEL, as a combination of edge computing and FL, its resource allocation may be the key solution to the scaling problem to achieve efficient and stable model training. In the future, the joint resource allocation of FEL will be affected by several emerging technologies. the low latency and high bandwidth of 6G networks will enhance the efficiency of inter-device collaboration, while the development of heterogeneous ED requires smarter resource scheduling algorithms. In addition, privacy computing, layered architecture, quantum computing and dynamic resource allocation techniques will further enhance the flexibility and security of FEL systems in the future. This paper involves multi-faceted resource allocation and optimization, including joint computation and communication resource allocation, joint data selection and communication resource allocation, joint device scheduling and resource allocation, and joint network topology and resource allocation. By optimizing the resource allocation, overall performance of the system can be improved and the effective use of resources can be achieved. In the concluding part of the study, the possible value of resource allocation strategies in FEL in the future is predicted.

**Acknowledgement:** We are grateful to our families and friends for their unwavering understanding and encouragement.

**Funding Statement:** This work was supported in part by the National Natural Science Foundation of China under Grant No. 61701197, in part by the National Key Research and Development Program of China under Grant No. 2021YFA1000500(4), and in part by the 111 Project under Grant No. B23008.

**Author Contributions:** Study conception and design: Jingbo Zhang, Qiong Wu, Pingyi Fan; data collection: Jingbo Zhang, Qiong Wu; analysis and interpretation of results: Jingbo Zhang, Qiong Wu, Qiang Fan; draft manuscript preparation: All authors. All authors reviewed the results and approved the final version of the manuscript.

**Availability of Data and Materials:** Not applicable.

**Ethics Approval:** Not applicable.

**Conflicts of Interest:** The author declares no conflicts of interest to report regarding the present study.